\documentclass[journal]{IEEEtran}

\usepackage{amsmath}
\usepackage{amssymb}
\usepackage{foreign}
\usepackage{graphicx}
\usepackage[pagebackref=true,breaklinks=true,colorlinks,bookmarks=false,urlcolor=black]{hyperref}
\usepackage{tabularx}
\usepackage{tikz}

\newcolumntype{H}{>{\setbox0=\hbox\bgroup}c<{\egroup}@{}}

\usepackage{xcolor}
\newcommand{\hl}[1]{#1}  

\usepackage{xspace}
\makeatletter
\DeclareRobustCommand\onedot{\futurelet\@let@token\@onedot}
\def\@onedot{\ifx\@let@token.\else.\null\fi\xspace}
\def\eg{\emph{e.g}\onedot} 
\def\ie{\emph{i.e}\onedot}

\makeatother

\ifCLASSINFOpdf
\else
\fi
\usepackage{amsmath}
\ifCLASSOPTIONcompsoc
 \usepackage[caption=false,font=normalsize,labelfont=sf,textfont=sf]{subfig}
\else
 \usepackage[caption=false,font=footnotesize]{subfig}
\fi
\begin{document}
%

\title{OpenPifPaf:\\ Composite Fields for Semantic Keypoint Detection and Spatio-Temporal Association}

%
%
%

\author{Sven~Kreiss,
        Lorenzo~Bertoni,
        Alexandre~Alahi
\thanks{Sven Kreiss, Lorenzo Bertoni and Alexandre Alahi are at the VITA lab at EPFL,
1015~Lausanne, Switzerland,
e-mail: sven.kreiss@epfl.ch}
}

\maketitle

\begin{abstract}
Many image-based perception tasks can be formulated as detecting, associating and tracking semantic keypoints, \textit{e.g.}, human body pose estimation and tracking. In this work, we present a general framework that jointly detects and forms spatio-temporal keypoint associations in a single stage, making this the first real-time pose detection and tracking algorithm.
We present a generic neural network architecture that uses Composite Fields to detect and construct a \emph{spatio-temporal pose} which is a single, connected
graph whose nodes are the semantic keypoints (\eg, a person's body joints)
in multiple frames.
For the temporal associations, we introduce the Temporal Composite Association Field (TCAF) which requires an extended network architecture and training method beyond previous Composite Fields.
Our experiments show competitive accuracy while being an order of magnitude faster on multiple publicly available datasets such as COCO, CrowdPose and
the PoseTrack 2017 and 2018 datasets.
We also show that our method generalizes to any class of semantic keypoints such as car and animal parts to provide
a holistic perception framework that is well suited for urban mobility such as self-driving cars and delivery robots.
\end{abstract}


%
\IEEEpeerreviewmaketitle

\section{Introduction}

The computer vision community has made tremendous progress in solving fine-grained perception tasks such as human body joints detection and tracking~\cite{andriluka14cvpr,lin2014microsoft}. We can cast these tasks as detecting, associating and tracking \textit{semantic keypoints}. Examples of semantic keypoints are ``left shoulders'', ``right knees'' or the ``left brake lights of vehicles'' as opposed to keypoints used in classical feature detectors that focus on the local geometry of the pixel intensities, like ``corners'' and ``edges''.
However, the performance of semantic keypoint tracking in live video sequences has been limited in accuracy and high in computational complexity
and prevented applications to the transportation domain with real-time requirements like self-driving cars
and last-mile delivery robots.
The majority of self-driving car accidents is caused by ``robotic'' driving where
the self-driving car conducts an allowed but unexpected stop and a human driver crashes
into the self-driving car~\cite{crow2017howsafeareselfdrivingcars}.
At their core, self-driving cars lack social intelligence.
They are blind to the body language of surrounding pedestrians when every person is
only perceived as a bounding box.
Current pose detection and tracking methods are neither fast enough nor robust enough to occlusions to be viable for self-driving cars.
Tracking human poses in real-time will enable
self-driving cars to develop a finer-grained understanding of pedestrian behavior
and with that a better conditioned reasoning for more natural driving.

\begin{figure}[t]
  \centering
  \includegraphics[width=1.0\linewidth,trim=7cm 4cm 9cm 7cm,clip]{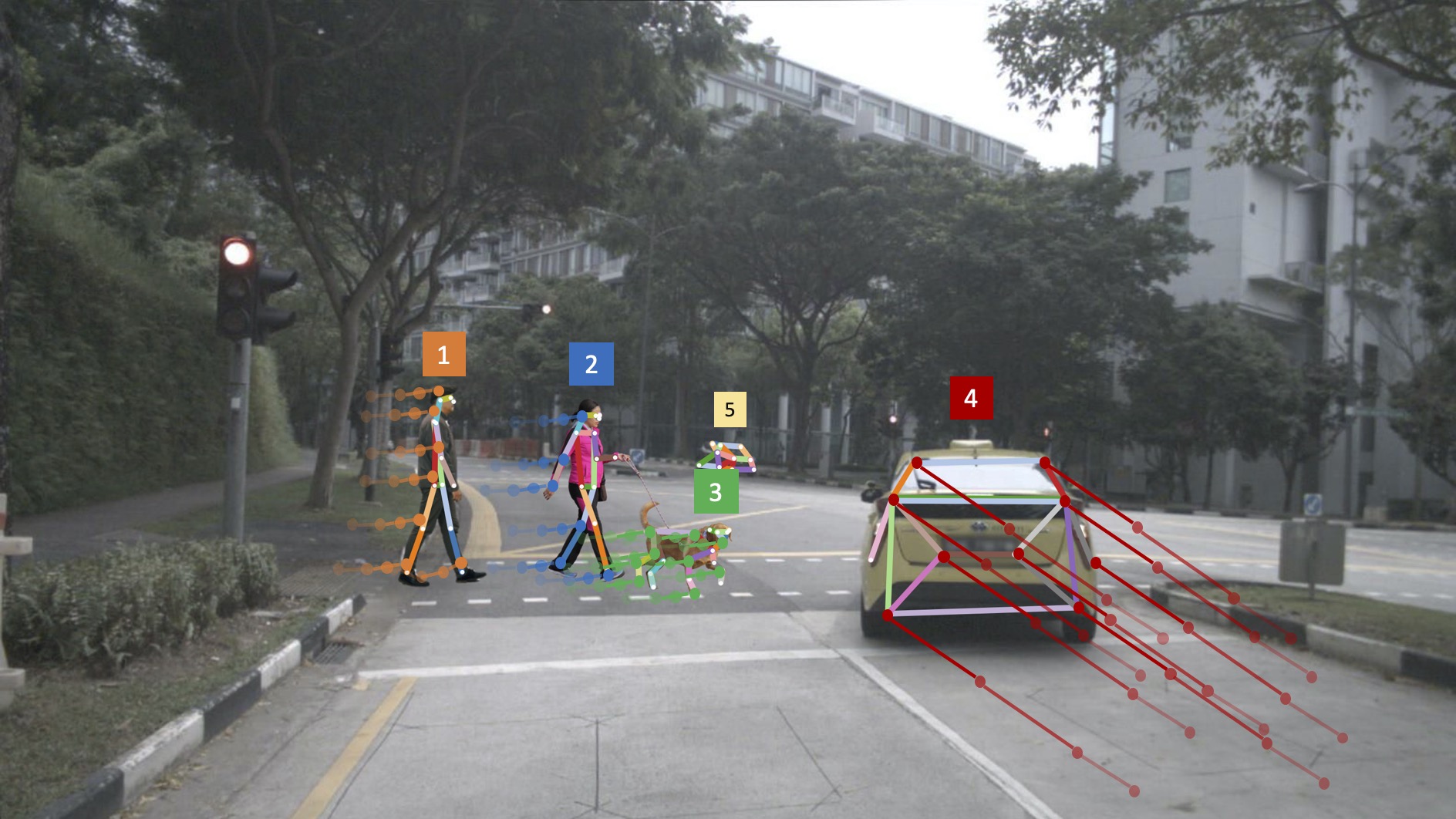}
  \caption{
    A real-world scene from the perspective of a self-driving car. Schematically, all moving
    actors are detected with their poses and tracked so that they can
    be consistently quantified over time. We place particular emphasis on
    understanding humans but also show generalizations to animals and cars.
    Here, a car (tracked as~4) is running a red light while also swerving to
    the right to avoid a woman (tracked as~2) who is walking her
    dog (tracked as~3).
  }
  \label{fig:pull}
\end{figure}

The problem is to estimate and track multiple human, car and animal poses in
image sequences, see Figure~\ref{fig:pull}.
The major challenges for tracking poses from the car perspective are
(i)~occlusions due to the viewing angle and
(ii)~prediction speed to be able to react to real-time changes in the environment.
Our method must be fast enough to be viable for
self-driving cars and robust to real-world variations like lighting, weather
and occlusions.

Although tracking has been studied extensively before human pose
estimation~\cite{milan2016mot16,kristan2015visual,kristan2017visual,lucas1981iterative},
a significant cornerstone that leverages poses are the works of
Insafutdinov \emph{et al.}~\cite{insafutdinov2017arttrack} and
Iqbal \emph{et al.}~\cite{iqbal2017posetrack} who pioneered multi-person pose tracking
for an arbitrary number of people in the wild. Both methods use graph matching
to track independent, single-frame poses over time.
To improve the matching for tracking, Doering \emph{et al.}~\cite{doering2018joint}
introduced temporal flow fields that improve the cost function for matching.
However, these works treat pose tracking as a multi-stage process:
infer single-frame poses -- which is itself a multi-stage process for top-down
methods -- and connect poses from frame to frame. This prohibits any improvement
to single-frame poses that could result from the temporal information
available in tracking. Here, we address these challenges by introducing a new method that jointly solves pose detection and tracking with Composite Fields.

First, we review Composite Fields for single-image multi-person pose estimation~\cite{kreiss2019pifpaf}.
Second, we introduce a new method for pose tracking. While single-frame pose estimation
can be viewed
as a pose completion task starting at a seed joint, we treat pose tracking
as a pose completion task starting with a pose from a previous frame and completing
a \emph{spatio-temporal pose}, which is a single, connected graph that spans space and time.
The spatio-temporal pose consists of at least two single-frame poses and additional
connections across the frames.

\begin{figure*}
  \centering
  \subfloat[]{\includegraphics[width=0.32\linewidth]{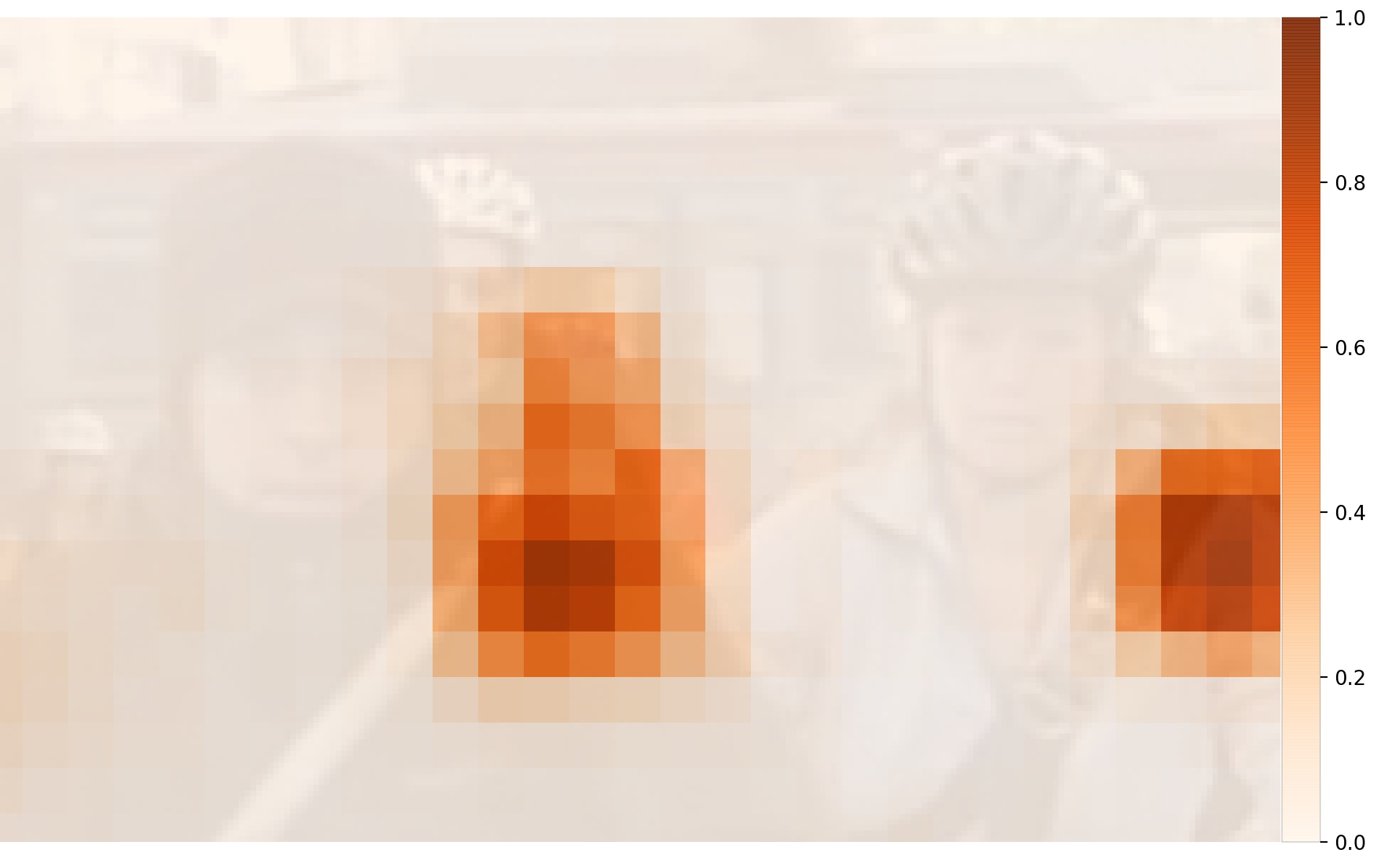}\label{fig:cif-confidence}}
  \hspace{0.1cm}
  \subfloat[]{\includegraphics[width=0.32\linewidth]{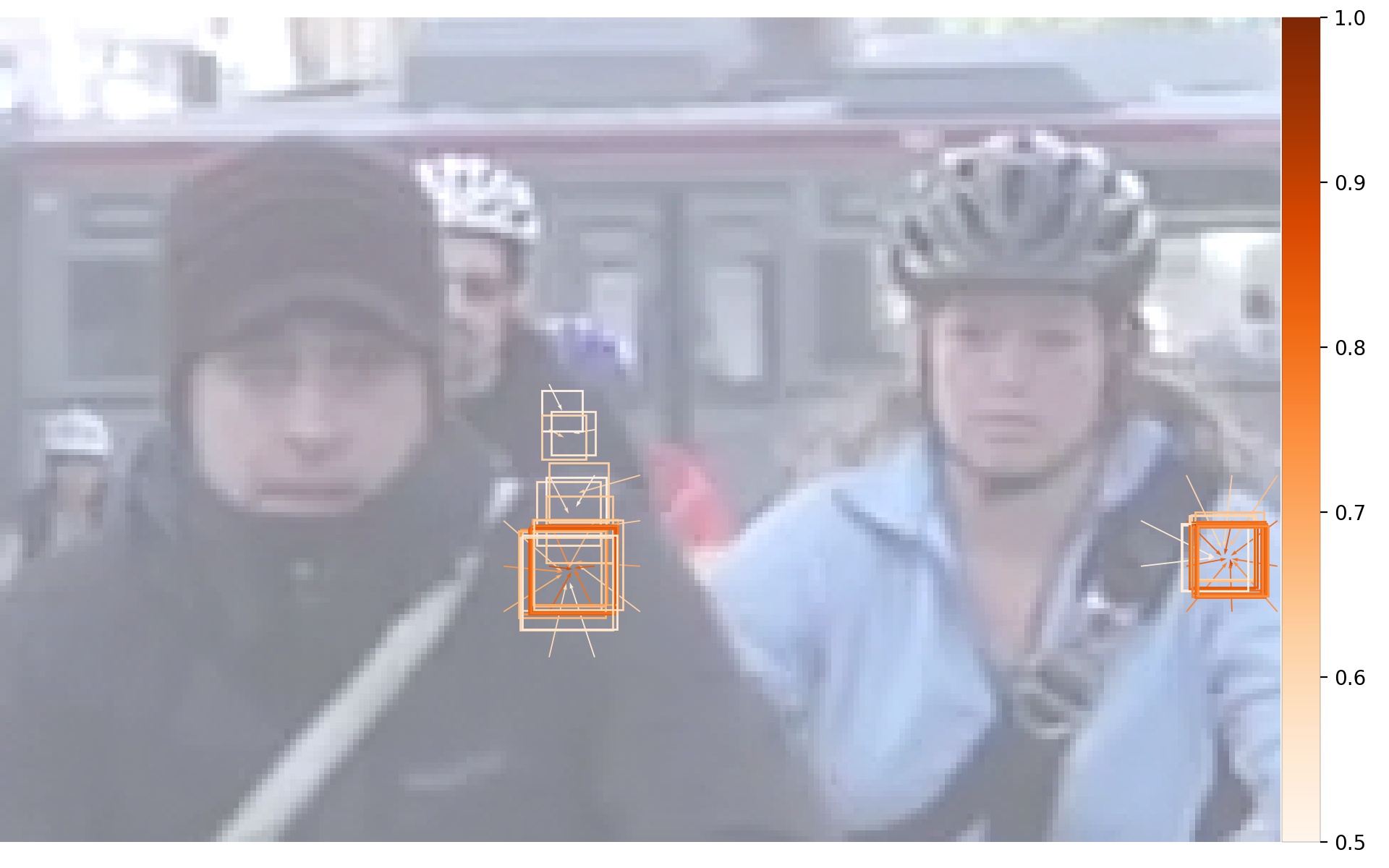}\label{fig:cif-regressions}}
  \hspace{0.1cm}
  \subfloat[]{\includegraphics[width=0.32\linewidth]{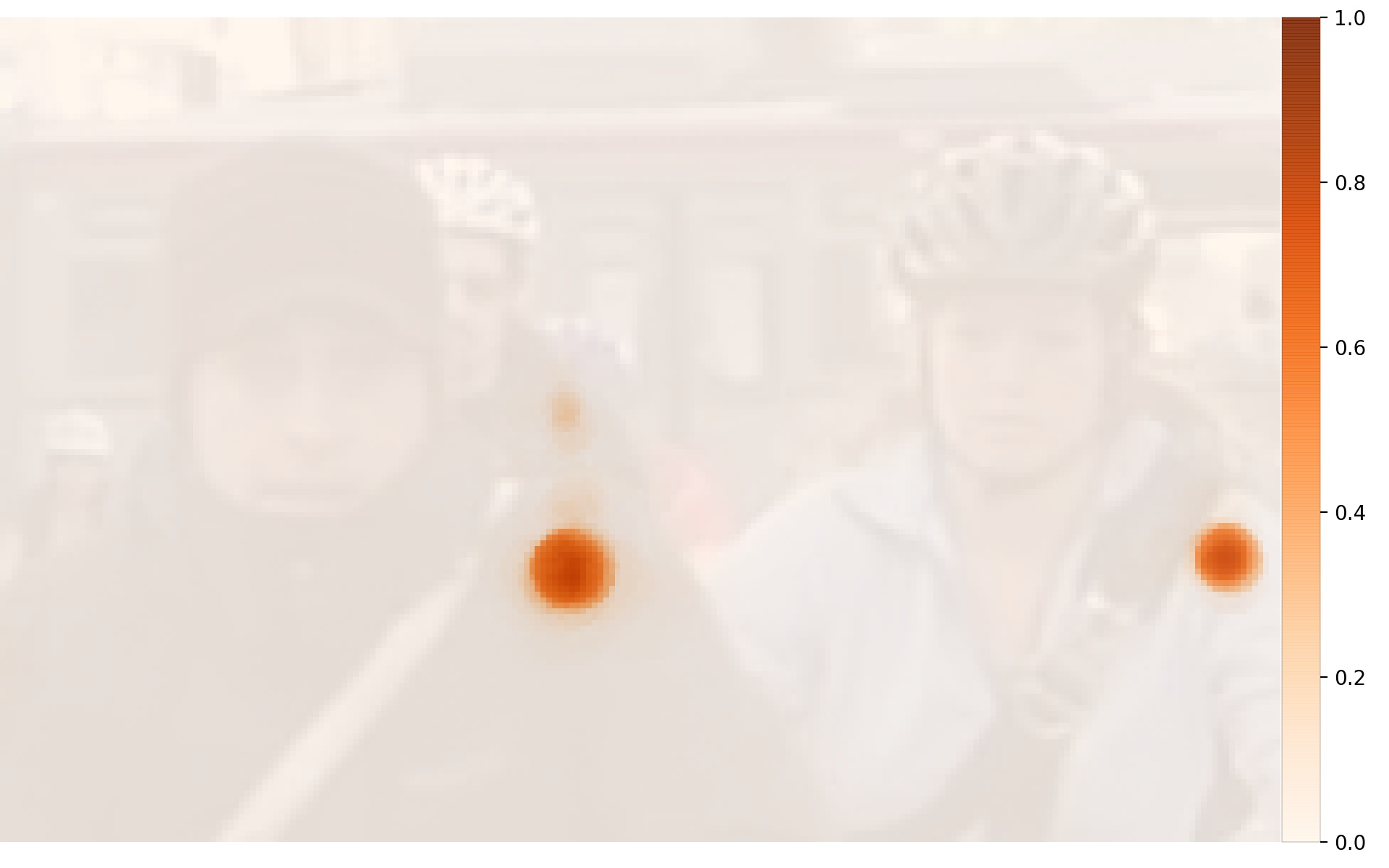}\label{fig:cifhr}}
  \caption{
    Visualizing the components of the CIF for the ``left shoulder'' keypoint
    on a small image crop.
    The confidence map is shown in~(\ref{fig:cif-confidence}).
    The vector field with joint-scale estimates is shown
    in~(\ref{fig:cif-regressions}). Only locations with
    confidence $> 0.5$ are drawn. The fused confidence, vector and scale
    components according to Equation~\ref{eq:cifhr} are shown
    in~(\ref{fig:cifhr}).
  }
  \label{fig:cif}
\end{figure*}

The contributions of this paper are
(i) a Temporal Composite Association Field (TCAF) which we use to form a
spatio-temporal pose and
(ii) a greedy decoder to jointly detect and track poses.
To the best of our knowledge, this method is the first single-stage, bottom-up pose detection and tracking method.
We outperform all previous methods in accuracy and speed
on the CrowdPose dataset~\cite{li2019crowdpose} with its particularly crowded images.
We perform
on par with the state-of-the-art bottom-up method for single-image human pose estimation on
the COCO~\cite{lin2014microsoft} keypoint task in precision and are an order of magnitude faster in speed.
Our model performs on par with the state-of-the-art method for human pose tracking on
PoseTrack~2017 and~2018~\cite{andriluka2018posetrack} while simultaneously being
an order of magnitude faster during prediction.
We also show that our method generalizes to car and animal poses which
demonstrates its suitability for a holistic perception framework.
Our method is implemented as an open source library, referred to as
\textit{OpenPifPaf}\footnote{\url{https://github.com/vita-epfl/openpifpaf_posetrack}}.

\section{Related Work}

\subsection{Pose Estimation}

State-of-the-art methods for pose estimation are based on
Convolutional Neural Networks~\cite{toshev2014deeppose,he2017mask,cao2017realtime,newell2017associative,papandreou2018personlab,xiao2018simple,sun2019deep,wei2016convolutional,newell2016stacked,kocabas2018multiposenet,cheng2020higherhrnet}.
All approaches for human pose estimation can be grouped into bottom-up and
top-down methods. The former estimates each body joint first and then groups
them to form a unique pose. The latter runs a person detector first and estimates
body joints within the detected bounding boxes.
Bottom-up methods were pioneered, \textit{e.g.}, by Pishchulin \emph{et al.} with
DeepCut~\cite{pishchulin2016deepcut}.
In their work, the part
association is solved with an integer linear program leading to processing
times for one image of the order of hours.
Newer methods use greedy decoders in combination with
additional tools to reduce prediction time as in
Part Affinity Fields~\cite{cao2017realtime},
Associative Embedding~\cite{newell2017associative},
PersonLab~\cite{papandreou2018personlab} and
multi-resolution networks with associate embedding~\cite{cheng2020higherhrnet}.
PifPaf~\cite{kreiss2019pifpaf} introduced composite fields for pose estimation
that produces a more precise association between joints than
OpenPose's Part Affinity Fields~\cite{cao2017realtime} and
PersonLab's mid-range fields~\cite{papandreou2018personlab}.
In the next section, we will review composite fields and show that they
generalize to tracking tasks.

\subsection{Pose Tracking}

Tracking algorithms can be grouped into top-down versus bottom-up approaches
for the pose part and the tracking part. Doering \emph{et al.}~\cite{doering2018joint}
were the first to introduce a method that is bottom-up in both the spatial and the temporal part.
They employ Part Affinity Fields~\cite{cao2017realtime} for the single-frame
poses in a Siamese architecture. The temporal flow fields (TFF) feed
into an edge cost computation for bipartite graph matching for tracking.
The idea is extended in MIPAL~\cite{hwang2019pose} for tracking limbs instead
of joints and in STAF~\cite{raaj2019efficient}.

Early work on multi-person pose tracking started
with~\cite{insafutdinov2017arttrack,iqbal2017posetrack}.
Recent work has shown excellent performance on the PoseTrack 2018 dataset
including the top-down method openSVAI~\cite{ning2018top} which decomposes
the problem into three independent stages of human candidate detection,
single-image human pose estimation and pose tracking. Similarly,
LightTrack~\cite{ning2020lighttrack} also builds a strong top-down pipeline
with interchangeable and independent modules.
Miracle~\cite{yu2018multi} uses a strong single-image pose estimator with
a cascaded pyramid network together with an IOU tracker.
HRNet for human pose estimation~\cite{sun2019deep} leverages a multi-resolution
backbone to produce high resolution feature maps that are context aware via
HRNet's multi-scale fusion.
In MSRA/FlowTrack~\cite{xiao2018simple}, optical flow is used to improve
top-down tracking of bounding boxes for tracking of human poses.
Pose-Guided Grouping (PGG)~\cite{jin2019multi} proposes a part
association method based on separate spatial and temporal embeddings.
KeyTrack~\cite{snower201915} uses pose tokenization and a transformer network to
associate poses.

\begin{figure*}
  \centering
  \subfloat[]{\includegraphics[width=0.49\linewidth]{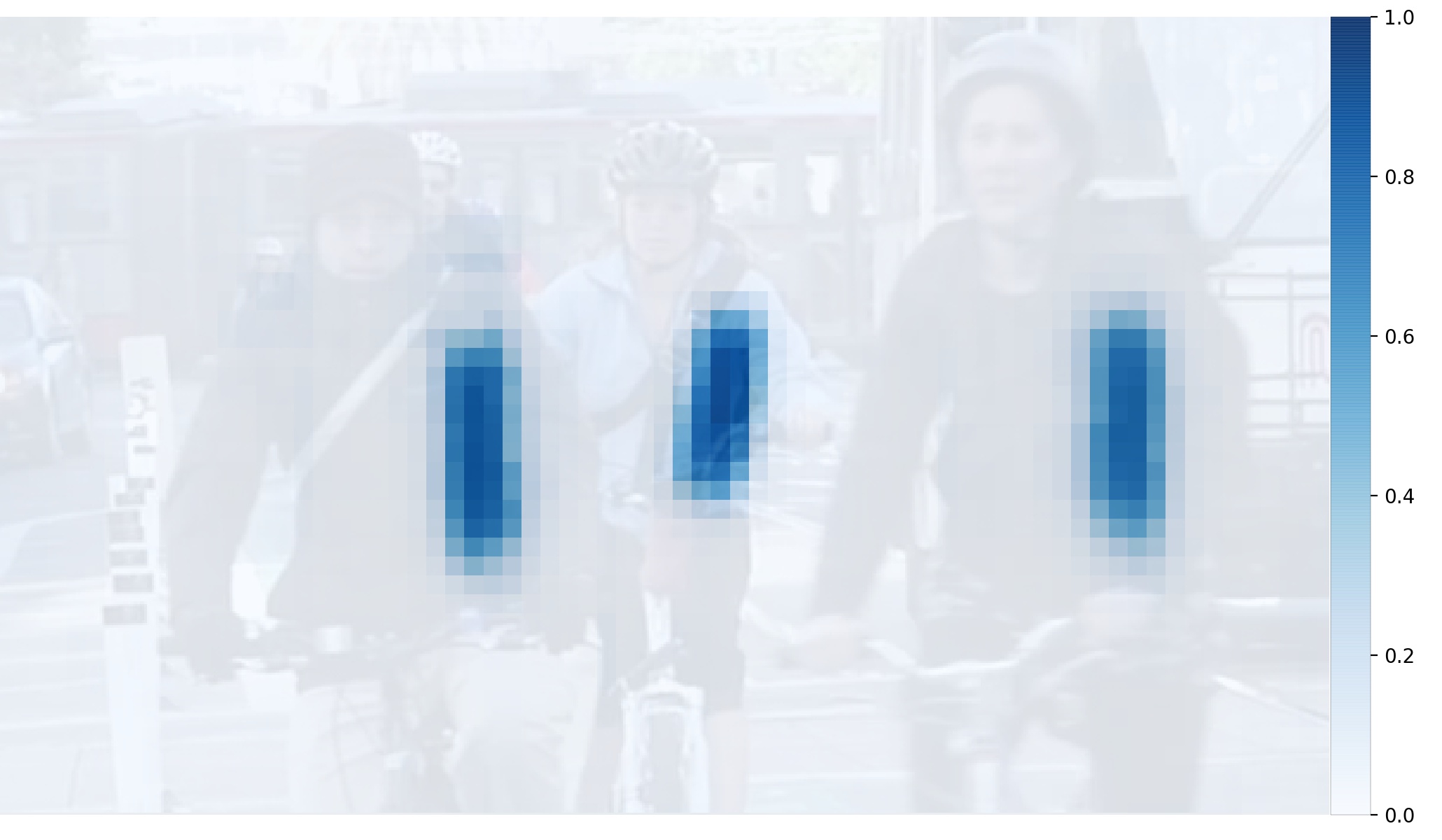}\label{fig:caf-intensity}}
  \hspace{0.1cm}
  \subfloat[]{\includegraphics[width=0.49\linewidth]{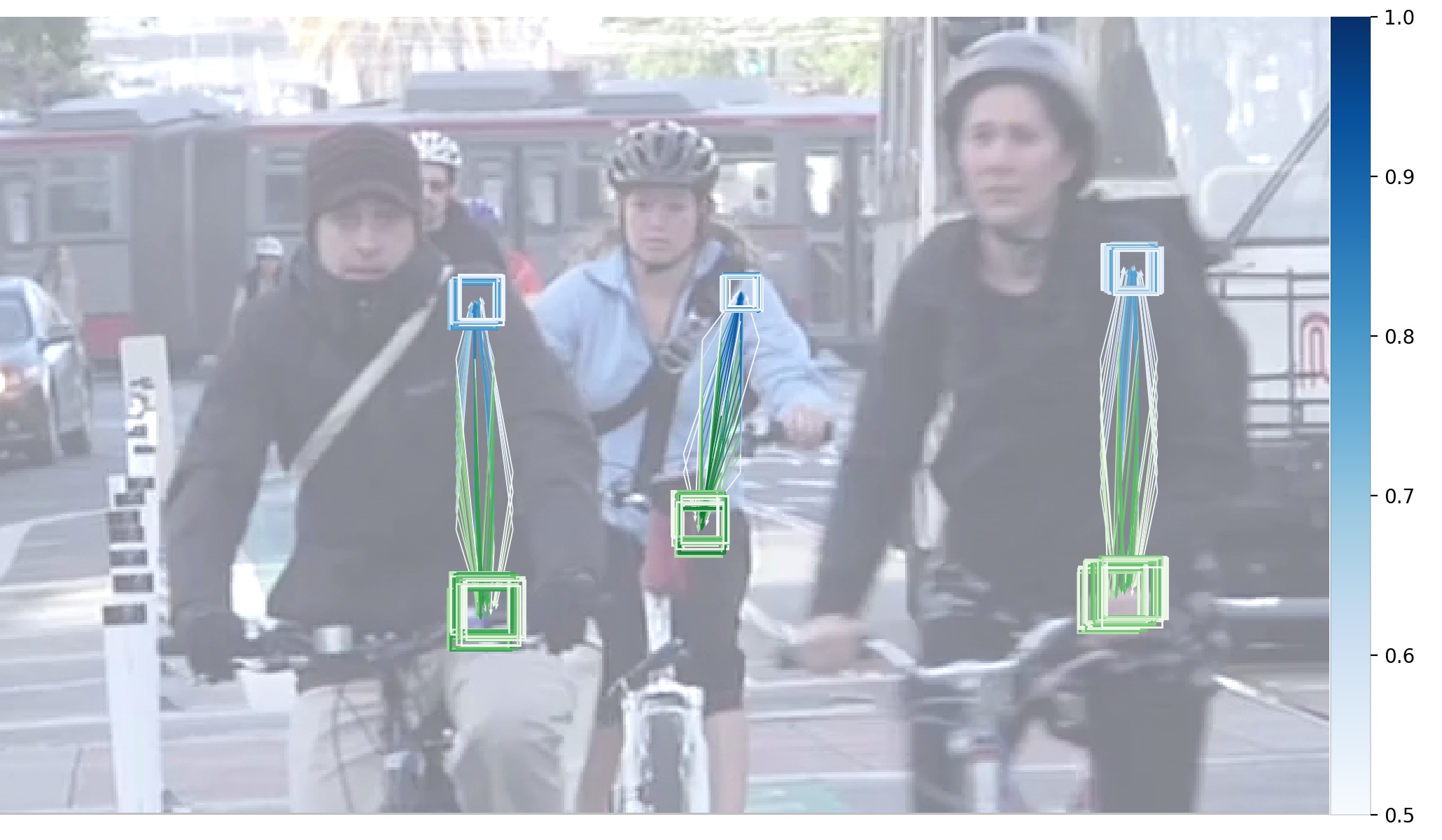}\label{fig:caf-regression}}
  \caption{
    Visualizing the components of the CAF that associates left shoulder with
    left hip. This is one of the 18 CAF. Every location of the feature map is
    the origin of two vectors which point to the shoulders and hips to associate.
    The confidence of associations $\textbf{a}_c$ is shown at their origin in
    (\ref{fig:caf-intensity}) and the vector components for
    $\textbf{a}_c$ greater than 0.5 are shown in (\ref{fig:caf-regression}).
  }
  \label{fig:caf-predicted}
\end{figure*}

\begin{figure}[t]
  \centering
  \subfloat[]{\includegraphics[height=2cm]{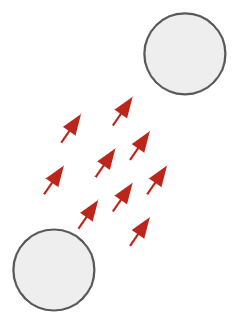}}
  \hspace{0.5cm}
  \subfloat[]{\includegraphics[height=2cm]{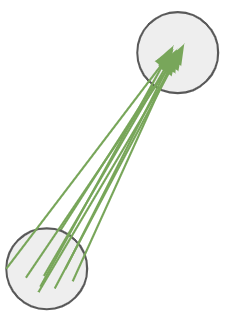}}
  \hspace{0.5cm}
  \subfloat[]{\includegraphics[height=2cm]{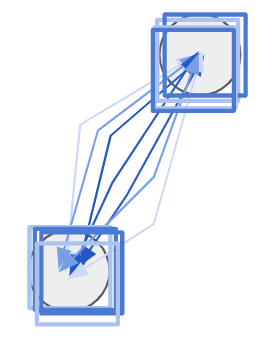}\label{fig:cartoon-caf}}
  \caption{
    Common association fields between two joints.
    Joints are visualized as gray circles.
    Part Affinity Fields~(a)
    as used in OpenPose~\cite{cao2017realtime}
    are unit vectors indicating a direction towards the next joint.
    Mid-range fields~(b) as used in PersonLab~\cite{papandreou2018personlab}
    are vectors originating
    in the vicinity of a source joint and point to the target joint.
    Our Composite Association Field~(c) regresses both source and target
    points and additionally predicts their joint size
    which are visualized with blue squares.
  }
  \label{fig:association-fields}
\end{figure}

\subsection{Beyond Humans}

While many state-of-the-art methods focused on human body pose detection and tracking, the research community has recently studied their performance on other classes such as animals and cars. Pose estimation research for animals and cars has to deal with additional challenges: limited labeled data~\cite{cao2019cross} and large number of self-occlusions~\cite{reddy2019occlusion}.

For animals, datasets are usually small and include limited animal
species~\cite{pereira2019fast, cao2019cross, biggs2020left,tigers2020, horses2021}.
To overcome this issue, DeepLabCut~\cite{mathis2018deeplabcut} and
WS-CDA~\cite{cao2019cross} have developed transfer learning techniques from
humans to animals. Mu \textit{et al.}~\cite{mu2020learning} have generated a
synthetic dataset from CAD animal models and proposed a technique to bridge the
real-synthetic domain gap. Another line of work has extended the human SMPL
model~\cite{loper2015smpl} to animals to learn simultaneously pose and shape of
endangered animals~\cite{zuffi2018lions, biggs2018creatures, zuffi2019three}.

For cars, self-occlusions between keypoints are inevitable. A few
methods improve performances by estimating 2D and 3D keypoints of vehicles
together. Occlusion-net~\cite{reddy2019occlusion} uses a 3D graph network with
self-supervision to predict 2D and 3D keypoints of vehicles using the CarFusion
dataset~\cite{dinesh2018carfusion}, while GSNet~\cite{ke2020gsnet} predicts 6DoF
car pose and reconstructs dense 3D shape simultaneously.  Without 3D information,
the popular OpenPose~\cite{cao2019openpose} shows qualitative results for vehicles
and Simple Baseline~\cite{sanchez2020simple} extends a top-down pose estimator for
cars on a custom dataset based on Pascal3D+~\cite{xiang2014beyond}.

\section{Composite Fields}

Our method relies on the \textit{Composite Fields} formalism to jointly detect and track \textit{semantic keypoints}. Hereafter, we briefly present them.

\paragraph{Field Notation}

Fields are functions over locations (\eg, feature map cells) and their outputs are primitives like scalars or
composites.
Composite Fields as introduced in~\cite{kreiss2019pifpaf} jointly
predict multiple variables of interest, for example, the
confidence, precise location and size of a semantic keypoint (\textit{e.g.}, body joint).

We will enumerate the spatial output coordinates of the neural network with
$i,j$ and reserve $x,y$ for real-valued coordinates in the input image.
A field over $(i, j)$ is denoted with $\textbf{f}^{ij}$ and can have
scalar, vector or composite values.
For example, the composite field of scalars $s$ and 2D vector components $v_x, v_y$
is $\{s, v_x, v_y\}^{ij}$. This is equivalent to ``overlaying'' a
confidence map with a vector field if the ground truth is aligned. This
equivalence is trivial in this example but becomes more subtle when we
discuss association fields below.

\begin{figure*}[t]
  \centering
  \includegraphics[width=1.0\linewidth,trim=0.5cm 0 0.3cm 0,clip]{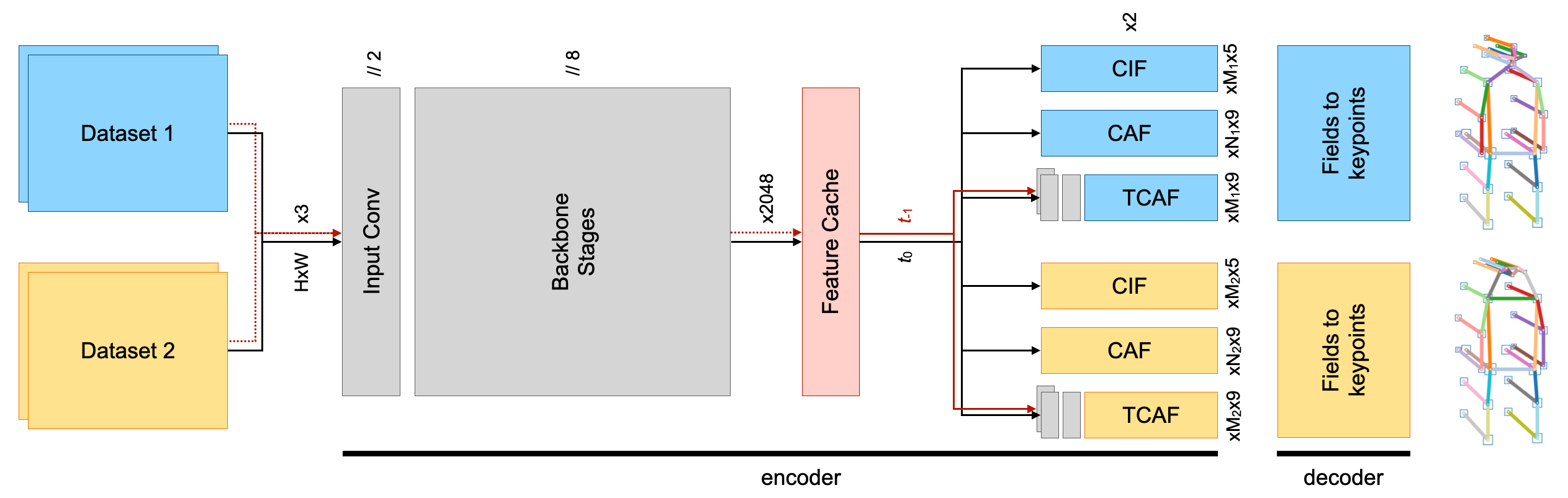}
  \caption[Model architecture]{
    Model architecture. The input is an image batch of size $(H,W)$ with three color
    channels, indicated by ``x3''.
    During joint training on multiple datasets, the datasets produce image pairs (black arrows for current image at $t_0$ and red arrows for image at $t_{-1}$) whereas during evaluation they produce single images in a sequence.
    The neural network based encoder produces composite fields
    for $M$ joints and $N$ connections.
    An operation with stride two is indicated by ``//2''.
    The shared backbone is a ResNet~\cite{he2016deep}
    or ShuffleNetV2~\cite{ma2018shufflenet} without max-pooling.
    The Feature Cache is only used during evaluation and injects for every image the previous
    feature map into the batch.
    We use a single $1 \times 1$ convolution in each head network.
    The TCAF head networks have a shared pre-processing step
    consisting of a feature reduction to 512 with a $1 \times 1$
    convolution followed by ReLU, a concatenation of the two feature maps
    and another $1 \times 1$ convolution with ReLU activation.
    For optional spatial upsampling, we append a sub-pixel convolution
    layer~\cite{shi2016real} to each head network.
    The decoder converts a set of composite fields into pose
    estimates. Each semantic keypoint is represented by a confidence score,
    a real-valued ($x$, $y$) coordinate pair and a size estimate.
  }
  \label{fig:model}
\end{figure*}

\paragraph{Composite Intensity Fields (CIF)}

The Composite Intensity Fields (CIF) characterize the intensity of semantic
keypoints. The composite structure is based on~\cite{papandreou2017towards}
with the extension of a scale $\sigma$ to characterize the keypoint
size. This is identical to the part intensity field in~\cite{kreiss2019pifpaf}.
We use the notation
$\textbf{p}^{ij}_J = \{c, x, y, b, \sigma\}^{ij}_J$ where $J$ is a particular
body joint type, $c$ is the confidence, $x$ and $y$ are regressed coordinates,
$b$ is the uncertainty in the location and $\sigma$ is the size of the joint.

Figure~\ref{fig:cif} shows the components of a CIF field and a high resolution
accumulation of the predicted intensity. The field is coarse with a stride of
16 with respect to the input image but the accumulated intensity is at high
resolution.
The high resolution confidence map $f(v, w)$ is a convolution
of an unnormalized Gaussian kernel $\mathcal{N}$ with width $\sigma$ over
the regressed targets from
the Composite Intensity Field $x$ and $y$ weighted by its confidence~$c$:
\begin{equation}
  \label{eq:cifhr}
  f_J(v, w) = \sum_{ij} c^{ij}_J \; \mathcal{N} (v, w | x^{ij}_J, y^{ij}_J, \sigma^{ij}_J)
\end{equation}
where $v$ and $w$ are real-valued coordinates in the image.
This accumulation incorporates information of the confidence $c$, the precisely
regressed spatial location $(x, y)$ and the predicted joint size $\sigma$.
This map $f_J$ is used to seed the pose decoder and to rescore predicted
CAF associations.

\paragraph{Composite Association Fields (CAF)}

Efficiently forming associations is the core challenge for
tracking multiple poses in a video sequence. The most difficult
cases are crowded scenes and camera angles where people occlude
other people -- as is the case in the self-driving car perspective
where pedestrians occlude other pedestrians.
Top-down methods first estimate bounding boxes and then do
single-person pose estimation per bounding box. This assumes
non-overlapping bounding boxes which is not given in our scenario.
Therefore, we focus on bottom-up methods.

In~\cite{kreiss2019pifpaf}, we introduced Part Association Fields
to connect joint locations together into poses.
Here, we extend this field with joint-scale components and call
it Composite Association Field~(CAF) to distinguish it better from
Part Affinity Fields introduced in~\cite{cao2017realtime}.
A graphical review of association fields is shown in
Figure~\ref{fig:association-fields} and shows that our CAF
expresses the most detail about an association.

CAFs predict a confidence, two vectors to the two parts this association is
connecting, two spreads $b$ for the spatial precisions of the regressions
(details in Section~\ref{sec:single-image}) and two joint sizes $\sigma$.
CAFs are represented with
$\textbf{a}^{ij}_{J_1 \leftrightarrow J_2} = \{c, x_1, y_1, x_2, y_2, b_1, b_2, \sigma_1, \sigma_2\}^{ij}_{J_1 \leftrightarrow J_2}$
where $J_1 \leftrightarrow J_2$ is the association between body joints $J_1$ and $J_2$.
Predicted associations between left shoulders
and left hips are shown for an example image in Figure~\ref{fig:caf-predicted}.
In our representation of an association, physically meaningful
quantities are regressed to continuous variables and do not
suffer from the discreteness of the feature map.
In addition, it is important to represent associations between two joints that are at the same pixel location. Our representation
is stable for these zero-distance associations -- something that
Part Affinity Fields~\cite{cao2017realtime} cannot do -- which
becomes particularly important when we introduce our extension for
tracking.

\section{Method}

We aim to present a method that can detect, associate and track \textit{semantic keypoints} in videos efficiently.
We place particular emphasis on urban and crowded scenes that are
difficult for autonomous vehicles.
Many previous methods struggle when object bounding boxes overlap.
In bird-eye views from drones or security cameras, bounding boxes are more
separated than in a car driver's perspective. Here, top-down methods struggle.
Previous bottom-up methods have been trailing top down methods in accuracy
without improving on performance either. Our bottom-up method is efficient,
employs a stable field representation and has high accuracy and
performance that even surpasses top-down methods.

Figure~\ref{fig:model} presents our model architecture. It is a shared
ResNet~\cite{he2016deep} or ShuffleNetV2~\cite{ma2018shufflenet} base
network without max-pooling.
The head networks are shallow and not shared between datasets.
In our examples, each dataset has a head network for joint intensities
(Composite Intensity Field -- CIF) and a head network for associations
(Composite Association Field -- CAF). Beyond CIF and CAF, additional head
networks can be added. In Section~\ref{sec:pose-tracking}, we
introduce the new Temporal Composite Association Field (TCAF) which is
predicted by an additional head network to facilitate pose tracking.

\begin{figure}
  \centering
    \includegraphics[height=4.8cm,trim=2cm 0 2.5cm 0,clip]{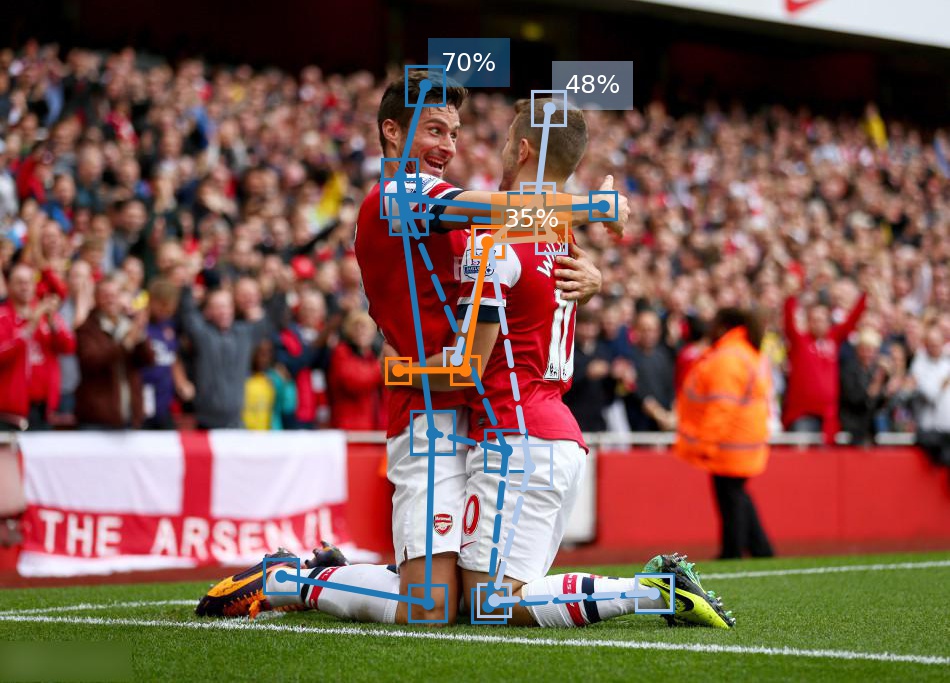}
    \includegraphics[height=4.8cm,trim=2cm 0 2.5cm 0,clip]{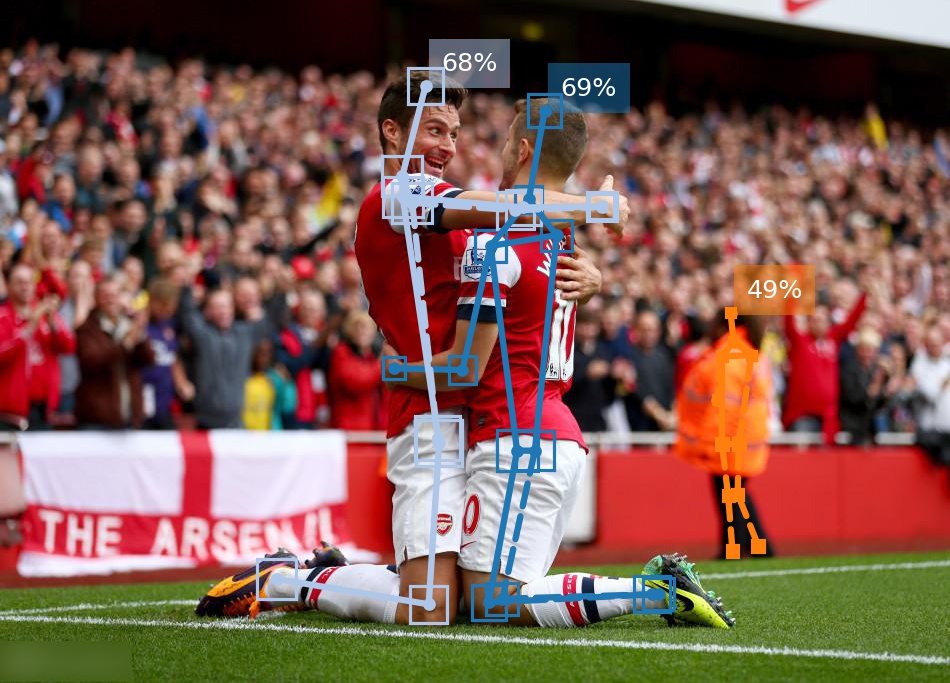}
  \caption{
    Effect of self-hidden keypoint suppression
    during training. The left image is without and the right image is with self-hidden keypoint suppression.
    The left hips of both soccer players collide in pixel space.
  }
  \label{fig:self-hidden-kp-suppression}
\end{figure}

We will introduce a tracking method that is a direct extension of single-image pose estimation. Therefore, we first
introduce our method for single-image pose estimation with
particular emphasis on details that will be relevant for pose
tracking.

\vspace{0.1cm}
\subsection{Single-Image Pose Estimation}
\label{sec:single-image}

\paragraph{Loss Functions for Composite Fields}
\label{sec:loss-functions}
Human pose estimation algorithms tend to struggle with the diversity of scales that
a human pose can have in an image. While a localization error for the joint of
a large person can be minor, that same absolute error might be a major mistake
for a small person. Our loss is the logarithm of the probability that all
components are ``well'' predicted, \ie, it is the sum of the log-probabilities
for the individual components. Each component follows standard loss prescriptions.
We use binary cross entropy (BCE) for classification with a Focal loss
modification $w$~\cite{lin2017focal}.
To regress locations in the image, we use the Laplace loss~\cite{kendall2017uncertainties}
which is an $L_1$-type loss
that is attenuated by a predicted spread $\hat{b}$ in the location.
To regress additional scale components (keypoint sizes), we use a Laplace loss with
a fixed spread $b_\sigma$ = 3.
The CIF loss function is:
\begin{eqnarray}
  \mathcal{L}_{\textrm{CIF}} &=
  & \sum\limits_{m_c}
      w(c,\hat{c}) \textrm{BCE}(c, \hat{c})
  \label{eq:loss-confidence} \\
  &&+ \;\; \sum\limits_{m_v} \frac{1}{\hat{b}} L_2(v, \hat{v}, b_\textrm{min})
    + \log \hat{b}
  \label{eq:loss-localization} \\
  &&+ \;\; \sum\limits_{m_s}
    \frac{1}{b_s} \left| 1 - \frac{\hat{s}}{s} \right|
  \label{eq:loss-scale}
\end{eqnarray}
with its three parts for confidence~(\ref{eq:loss-confidence}),
localization~(\ref{eq:loss-localization}) and scale~(\ref{eq:loss-scale})
and where:
\begin{equation}
  L_2(v, \hat{v}, b_\textrm{min}) = \sqrt{(v_1 - \hat{v}_1)^2 + (v_2 - \hat{v}_2)^2 + b_\textrm{min}^2} \;\;\;.
\end{equation}
The sums are over masked feature cells $m_c$, $m_v$ and $m_\sigma$ with $i,j,J$ implied.
The mask for confidence $m_c$ is almost the entire image apart from regions
annotated as ``crowd regions''~\cite{lin2014microsoft}. The masks for localization
$m_v$ and for scale $m_\sigma$ are only active in a $4\times4$ window around the
ground truth keypoint. Per feature map cell, there is a ground truth confidence
$c$ and its predicted counterpart $\hat{c}$. The predicted location
$\hat{v} = (\hat{v}_1, \hat{v}_2)$ is optimized with a Laplace loss with a predicted
spread $\hat{b}$ for heteroscedastic aleatoric
uncertainty~\cite{kendall2017uncertainties} with respect to the ground truth
location~$v$.
A $b_\textrm{min} = 1\textrm{px}$ is added to prevent exploding losses when
the spread becomes too small. For stability, we clip the BCE loss when it becomes
larger than five.
The CAF loss has the same structure but with two localization
components~(\ref{eq:loss-localization}) and two scale components~(\ref{eq:loss-scale}).

\paragraph{Self-Hidden Keypoint Suppression}

The COCO evaluation metric treats visible and hidden keypoints
in the same manner. As in~\cite{kreiss2019pifpaf}, we include
hidden keypoints in our training. However, when a visible
and a hidden keypoint appear close together, we remove the hidden
keypoint from the ground truth annotation so that this keypoint
is not included in associations.
In Figure~\ref{fig:self-hidden-kp-suppression}, we show the
effect of excluding these self-hidden keypoints from training
and observe better pose reconstruction when a keypoint hides
another keypoint of the same type.

\paragraph{Greedy Decoder with Frontier}

The composite fields are converted into sets of pose estimates
with the greedy decoder introduced in~\cite{kreiss2019pifpaf} and reviewed here.
The CIF field and its high-resolution accumulation $f(x,y)$
defined in equation~\ref{eq:cifhr} provide seed locations.
Previously, new associations were formed starting at the joint
that has currently the highest score without taking
the CAF confidence of the association into account.
Here, we introduce a frontier which is a priority queue of
possible next associations.
The frontier is ordered by the possible future joint scores which are a function
of the previous joint score and the best CAF association:
\begin{equation}
  \max_{ij} s(\textbf{a}^{ij}_{J_1 \leftrightarrow J_2}, \vec{x}) = c \exp\left(-\frac{||\vec{x} - (x_1, y_1)||_2}{\sigma_1}\right) f_{J_2}(x_2, y_2)
\end{equation}
where $\vec{x}$ is the source joint location, $\textbf{a}^{ij}_{J_1 \leftrightarrow J_2} = (c, x_1, y_1, x_2, y_2, \sigma_1, \sigma_2)$ is the CAF field with implied sub-/superscripts on the components
and $f_{J_2}$ is the high resolution confidence map of the target joint $J_2$.
An association is rejected when it fails reverse matching.
To reduce jitter, we not only use the best CAF association in the above equation but a weighted mixture of the best two associations; similar to blended connections
in~\cite{bazarevsky2019blazeface}.
Only when
all possible associations are added to the frontier,
the connection is made to the highest priority in the frontier.
This algorithm is fast and greedy. Once a connection to a new
joint has been made, this decision is final.

\paragraph{Instance Score and Non-Maximum Suppression (NMS)}
Once all poses are reconstructed, we apply NMS. Poses are first sorted by
their instance score which is the weighted mean of the keypoint scores where the three
highest keypoint scores are weighted three times higher.
We run NMS at the keypoint level as in~\cite{kreiss2019pifpaf,papandreou2018personlab}.
The suppression radius is dynamic and based on the predicted joint size.
We do not refine predictions.

\paragraph{Denser Pose Skeletons}

\begin{figure}[t]
  \centering
    \includegraphics[height=2.5cm,trim=0 6cm 0 6cm,clip]{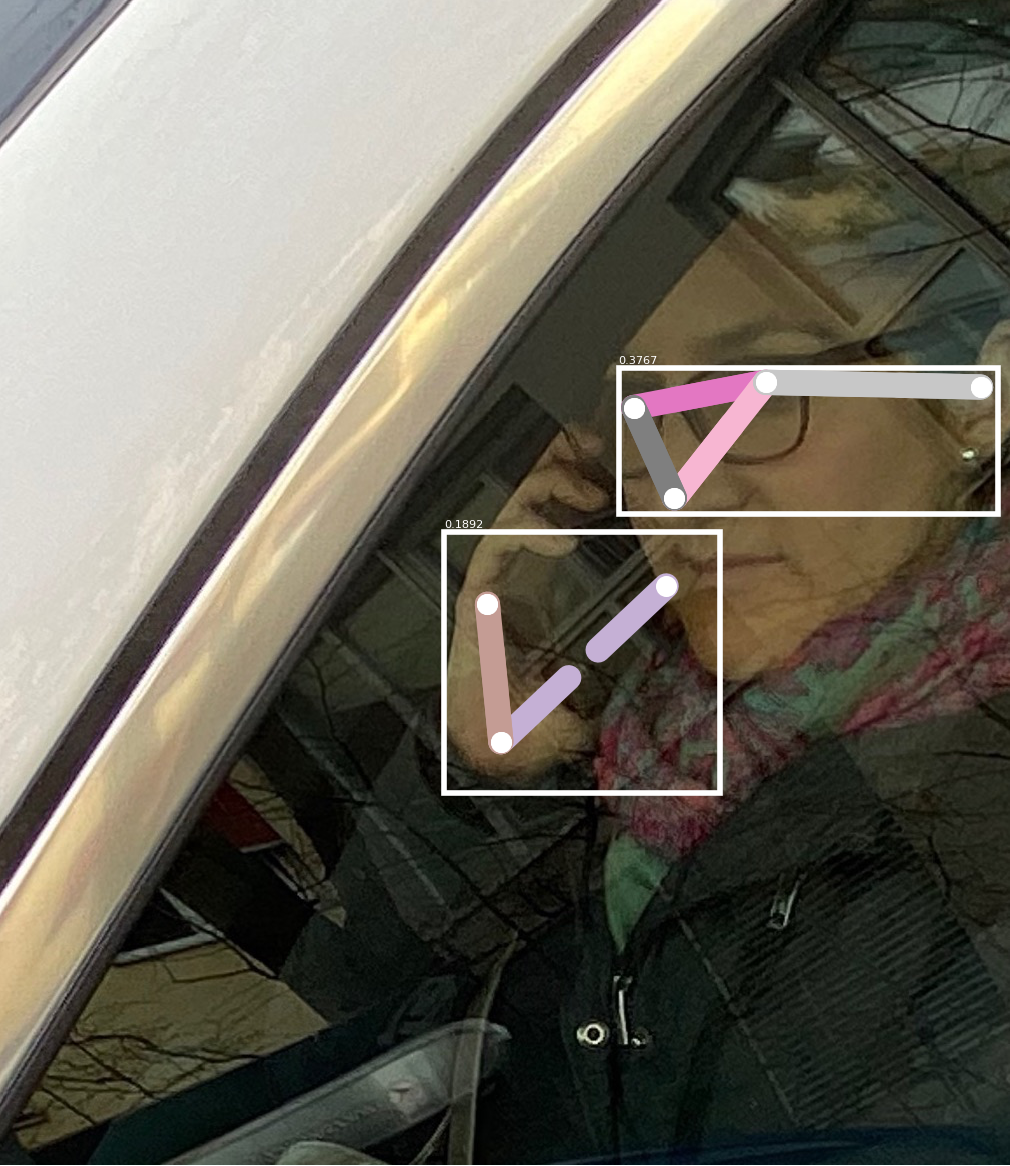}
    \includegraphics[height=2.5cm,trim=0 6cm 0 6cm,clip]{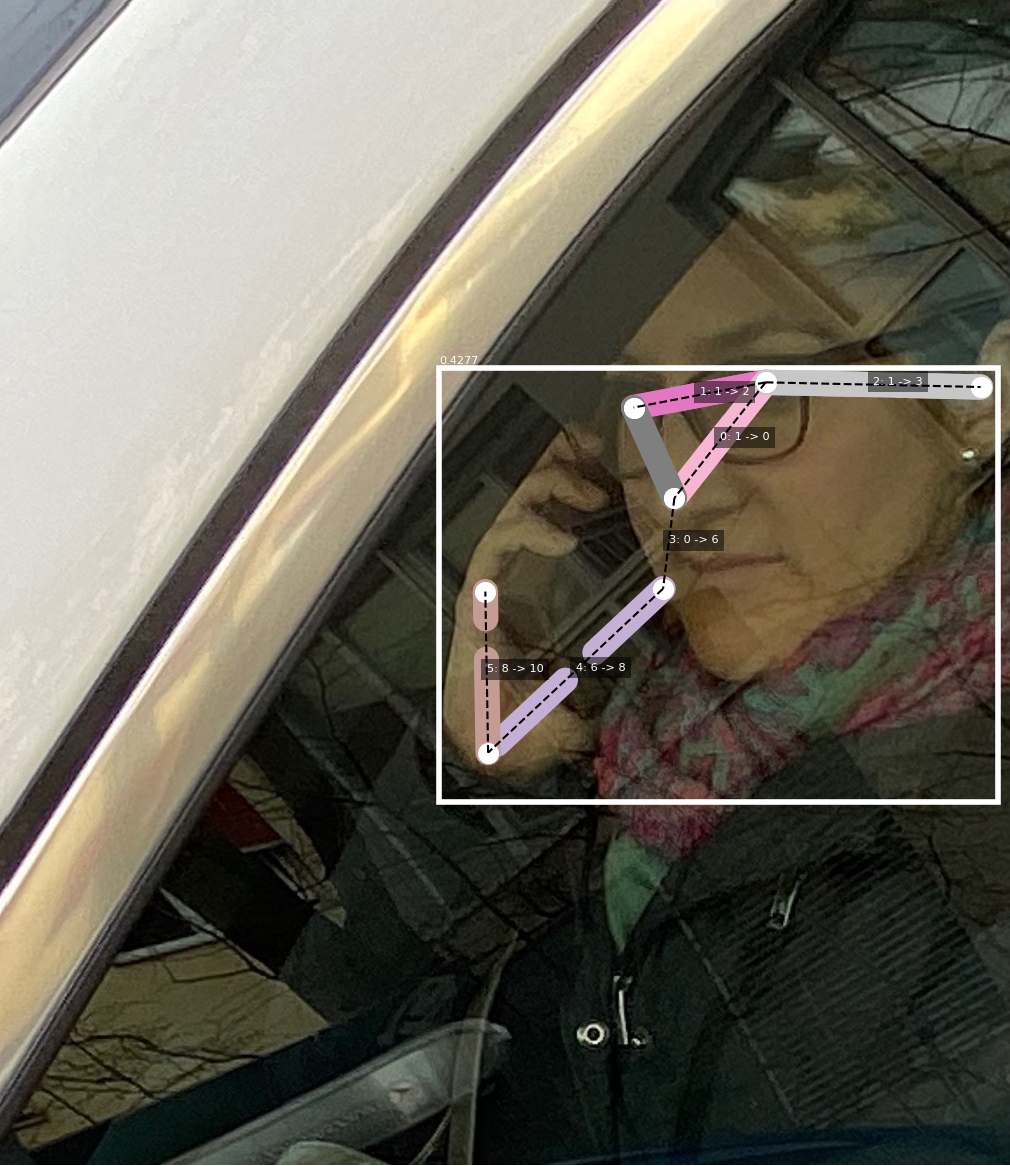}
  \caption{
    \emph{Left:} A sparse pose
    cannot connect the right arm to the facial keypoints leading to the detection
    of two separate person instances highlighted by the two white bounding boxes.
    \emph{Right:}
    An additional dense connection between the nose and right shoulder leads
    to a correctly identified single pose.
  }
  \label{fig:qual-dense}
\end{figure}

Figure~\ref{fig:poses2} gives an overview of the pose skeletons that are used
in this paper.
In particular, Figure~\ref{fig:poses2}b shows a modification of the standard COCO
pose~\cite{lin2014microsoft} with additional associations. These denser associations
are redundancies
in case of occlusions.
The additional associations are longer-range and therefore harder to predict.
The frontier in our greedy decoder takes
this difficulty into account and automatically prefers easier, confident
associations when available.
Qualitatively, the advantage of dense associations is shown in
Figure~\ref{fig:qual-dense}. With the standard COCO skeleton, the single
person's pose skeleton would be divided into two disconnected parts (left image)
as indicated by the two white bounding boxes.
With the additional denser associations, a single pose is formed (right image).

\subsection{Pose Tracking}
\label{sec:pose-tracking}

In the previous section we introduced our method for bottom-up pose estimation
in single images.
We now
generalize that method to tracking poses in videos with associations between images in the
same bottom-up fashion. Our unified approach forms both spatial and
temporal associations simultaneously. This even leads to improved single-image
poses from the additional temporal information.

\paragraph{Temporal Composite Association Field (TCAF)}

During training, tracking data is fed into the base network as image pairs
that are concatenated
in the batch dimension, \ie, a batched input tensor of eight image pairs has
the same shape as 16 individual images.

During prediction, the backbone processes one image at a time and each image only once.
The resulting feature map is
then concatenated with the previous feature map from the ``Feature Cache''
(see Figure~\ref{fig:model}).
While there is still duplicate computation in the head networks, their
computational complexity is small.

To form
associations in image sequences, we introduce the Temporal Composite
Association Field (TCAF).
Its output structure is identical to a CAF
field, but its input is based on pairs of feature maps that were created
independently.
To jointly process information from both feature maps, the TCAF head
contains a preprocessing step of a $1 \times 1$ input convolution to reduce the feature size to 512 with ReLU non-linearity, a concatenation of these two feature maps to 1024 features, a $1 \times 1$ convolution with ReLU to process the two images jointly and a final $1 \times 1$ convolution to produce all components necessary for a composite association field.

\begin{figure}[t]
  \centering
  \subfloat[]{\includegraphics[height=2.7cm]{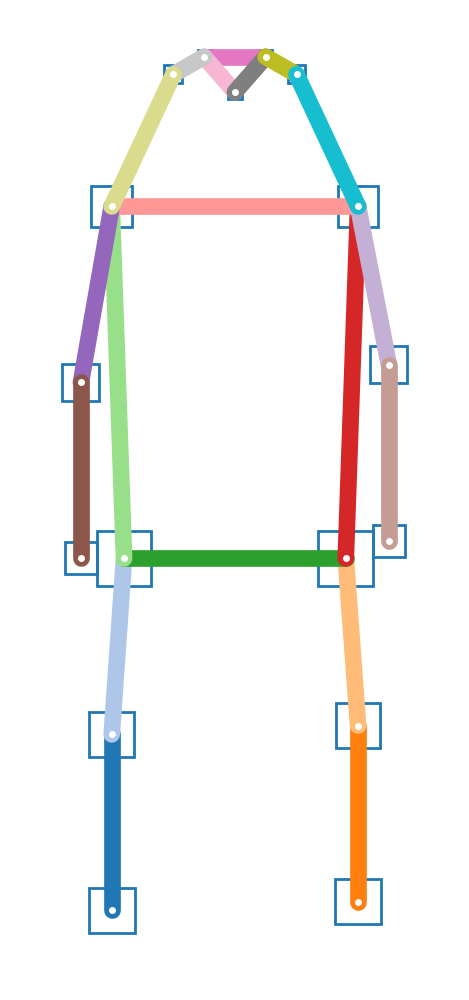}} 
  \subfloat[]{\includegraphics[height=2.7cm]{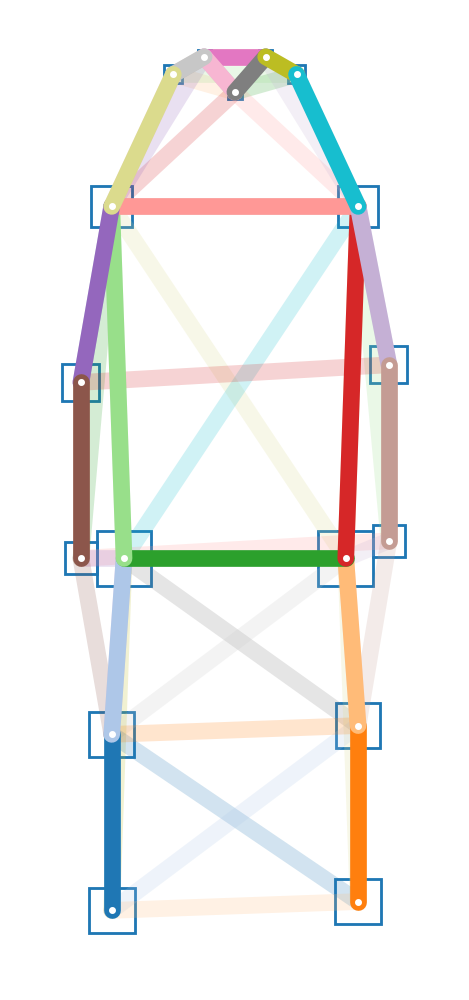}}  
  \subfloat[]{\includegraphics[height=2.4cm,trim=0 0.5cm 0 0.5cm]{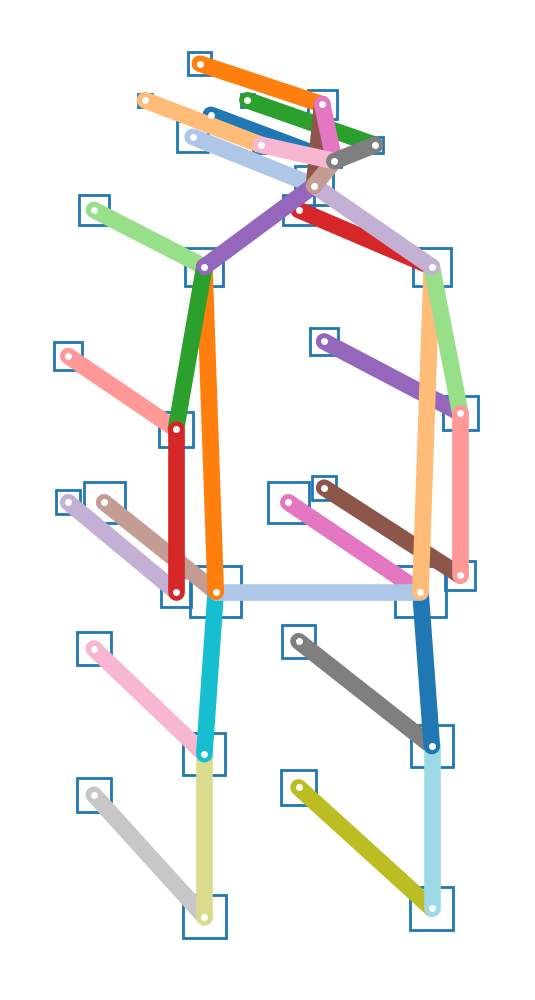}}  
  \subfloat[]{\includegraphics[height=2.8cm,trim=0.8cm 0.8cm 2.5cm 1cm,clip]{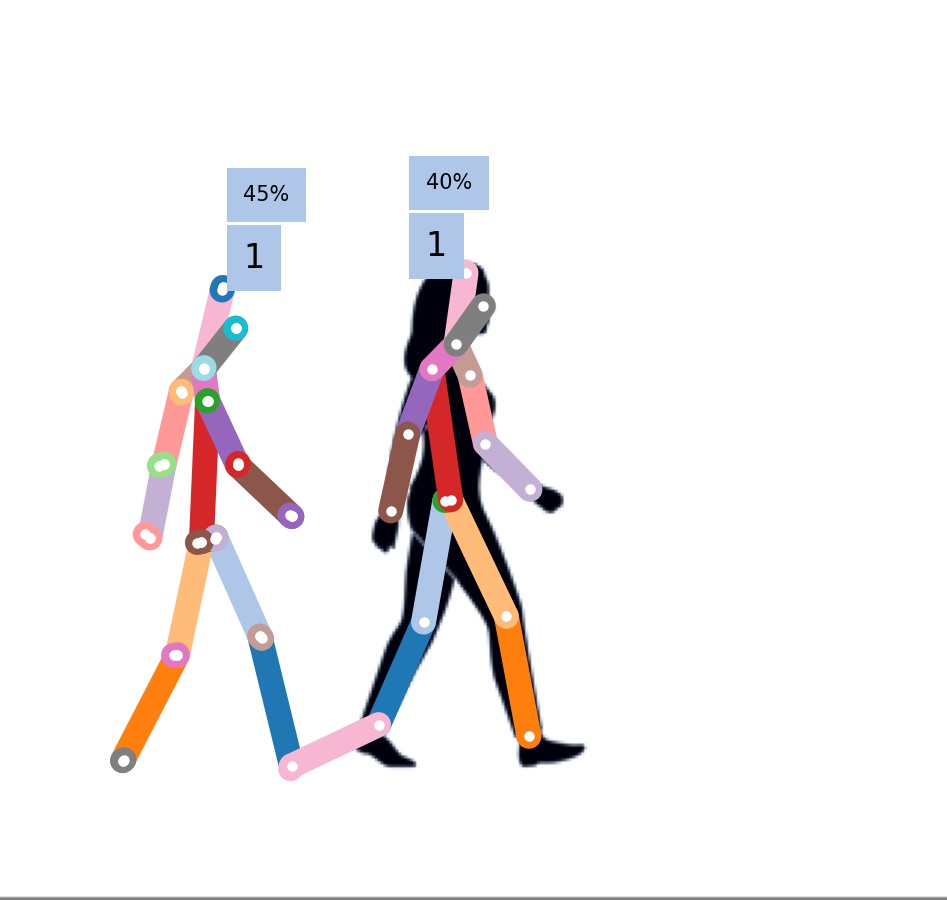}}
  \caption{
    A COCO person pose~\cite{lin2014microsoft} is shown in (a).
    Additional denser connections are shown in lighter colors in (b).
    The additional connections provide redundancies in case of occlusions.
    A pose skeleton as used in Posetrack with temporal connections
    is shown in (c).
    An example of a tracked pose is shown in (d). The first frame is captured with the right leg (blue) in front and the second frame one step later.
    For clarity, only connections that were used to decode the pose are shown and therefore only the temporal
    connection that is connecting the right ankle from the past frame to the current frame is visible.
  }
  \label{fig:poses2}
\end{figure}

\paragraph{Spatio-Temporal Poses}
\label{sec:st-pose}

Figure~\ref{fig:poses2}c shows a schematic of a person pose
(17 joints and 18 associations) with additional temporal
connections to all joints of the same kind in the previous frame.
In our method, this is treated as a single pose with $2\times17$ joints (CIF)
and 18 associations (CAF) within the same frame and an
additional 17 associations (TCAF) between frames.

\paragraph{From Spatio-Temporal Poses to Tracks}

Spatio-temporal poses create temporal associations in pairs of images. We now
introduce our book-keeping method to go from pairs of images to image sequences.
During evaluation and for a new frame $t_0$, the decoder creates new tracking
poses from existing tracks (poses in the previous frame $t_{-1}$) or from
single-image seeds in the current frame $t_0$.
These partial poses are then completed using the same greedy frontier decoder
described for single images.
Once all spatio-temporal poses are complete, the $t_0$ joints are extracted
into single-frame poses. Every single-frame pose is already tagged with an
existing track-id if the spatio-temporal pose was generated from an existing track
or a new track-id if the spatio-temporal pose originated from a new seed in the current frame.
The single-frame poses are
then filtered with soft NMS~\cite{papandreou2018personlab} and then either
added to existing tracks or they become the first poses of new tracks.

Our method is bottom-up in both pose estimation and
tracking and estimates temporal and spatial connections within a
single stage. Most existing work -- even other bottom-up
tracking methods~\cite{doering2018joint,hwang2019pose} -- employ a two
stage process where, first, spatial connections
are estimated and, second, temporal connections are made.

\section{Experiments}

Self-driving cars must perceive and predict pedestrians and other traffic
participants robustly.
One of the most challenging scenarios are crowded places.
We will first show experiments on single-image human pose estimation in
CrowdPose~\cite{li2019crowdpose} which contains particularly challenging
scenarios and on the standardized and competitive COCO~\cite{lin2014microsoft}
person keypoint benchmark. Then we will show results for pose tracking in videos
on the PoseTrack~2017~\cite{iqbal2017posetrack}
and~2018~\cite{andriluka2018posetrack} datasets.
We have conducted extensive experiments to show the benefit of a
unified bottom-up pose estimation and tracking method with spatio-temporal poses.
To demonstrate the universality of our approach, we apply our method also
to poses of cars and poses of animals.

\subsection{Datasets}

\paragraph{CrowdPose}
In~\cite{li2019crowdpose}, the CrowdPose dataset is proposed. It is a selection
of images from other datasets with a particular emphasis on how
crowded the images are. The crowd-index of an image represents the amount of
overlap between person bounding boxes. The authors place particular emphasis
on a uniform distribution of the crowd-index in all data partitions.
Because this dataset is a composition of other datasets and to avoid contamination,
our CrowdPose models are pretrained on ImageNet~\cite{deng2009imagenet} and
then trained on CrowdPose only.
The dataset comes with a split of 10,000 images for training, 2,000 for validation and
8,000 images for the test set.

\paragraph{COCO}
The de-facto standard for person keypoint prediction is the competitive COCO keypoint
task~\cite{lin2014microsoft}. The test set is private and powers an active
leaderboard via a protected challenge server.
COCO contains 56,599 diverse training images with person keypoint annotations.
The validation and test-dev sets contain 5,000 and 20,288 images.

\paragraph{ApolloCar3D}
We generalize our approach to vehicle keypoints using the ApolloCar3D
dataset~\cite{song2019apollocar3d}, which contains 5,277 driving images at a
resolution of 4K and over 60K car instances. The authors defined 66 semantic
keypoints in the dataset and, for each car, they provided annotations for the
visible ones. For clarity, we choose a subset of 24 semantic
keypoints and show quantitative and qualitative results on this dataset.

\paragraph{Animal Dataset}
We evaluate the performances of our algorithm on the Animal-Pose
Dataset~\cite{cao2019cross}, which provides annotations for five categories of
animals: dog, cat, cow, horse, sheep for a total of 20 keypoints. The dataset
includes 5,517 instances in more than 3,000 images. The majority of these
images originally belong to the VOC dataset~\cite{everingham2015pascal}.

\paragraph{PoseTrack 2017 and 2018}
We conduct quantitative studies of our tracking performance on the
PoseTrack 2017~\cite{iqbal2017posetrack} and 2018~\cite{andriluka2018posetrack} datasets.
The datasets contain short video sequences of annotated and tracked human poses
in diverse situations. The PoseTrack 2018 dataset contains 593 training scenes,
170 validation scenes
and 375 test scenes. The test labels are private.
PoseTrack 2017 is a subset of the 2018 dataset with 292 train, 50 validation and
208 test scenes. However, the 2018 leaderboard is
frozen and new results are only updated for the 2017 leaderboard. Therefore, many
recent methods present results on the older, smaller dataset. Here, we will report
numbers for both 2017 and 2018.

\subsection{Evaluation}

\paragraph{Single-Image Multi-Person Poses}
Both CrowdPose and COCO follow COCO's keypoint evaluation method.
The object keypoint similarity~(OKS) score~\cite{lin2014microsoft} is used
to assign a bounding box to each keypoint as a function of the person instance
bounding box area. Similar to detection, the metric computes overlaps between ground truth
and predicted bounding boxes to compute the standard detection metrics
average precision~(AP) and average recall~(AR).

CrowdPose breaks down the test set at the image level into easy, medium and hard.
The easy set
contains images with a crowd index in $[0, 0.1]$, the medium set in $[0.1, 0.8]$
and the hard set in $[0.8, 1.0]$. Given the uniform crowd-index distribution,
most images of the test set are in the medium category.

COCO breaks down the precision scores at the instance level for medium
instances with a bounding
box area of (32~px)$^2$ to (96~px)$^2$ and for large instances with a bounding
box area larger than (96~px)$^2$.
For each image, pose estimators have to provide the 17 keypoint locations per pose
and a total score for each pose. Only the top 20 scoring poses per image
are considered for evaluation.

\paragraph{Pose Tracks}
A common metric to evaluate the tracking of human poses is the Multi Object
Tracker Accuracy (MOTA)~\cite{bernardin2008evaluating,milan2016mot16} which
is also the main metric in PoseTrack challenges and leaderboards.
It combines false positives, false negatives and ID switches into a single
metric.
We compare against the best methods that submitted to the
PoseTrack~2017 and~2018 evaluation server which computes all metrics on private
test sets. These methods include strong top-down methods as well as
bottom-up methods for pose estimation and tracking.

\begin{figure*}
  \centering
  \begin{tabular}{c}
    \includegraphics[height=5.1cm,trim=0 0 0 0,clip]{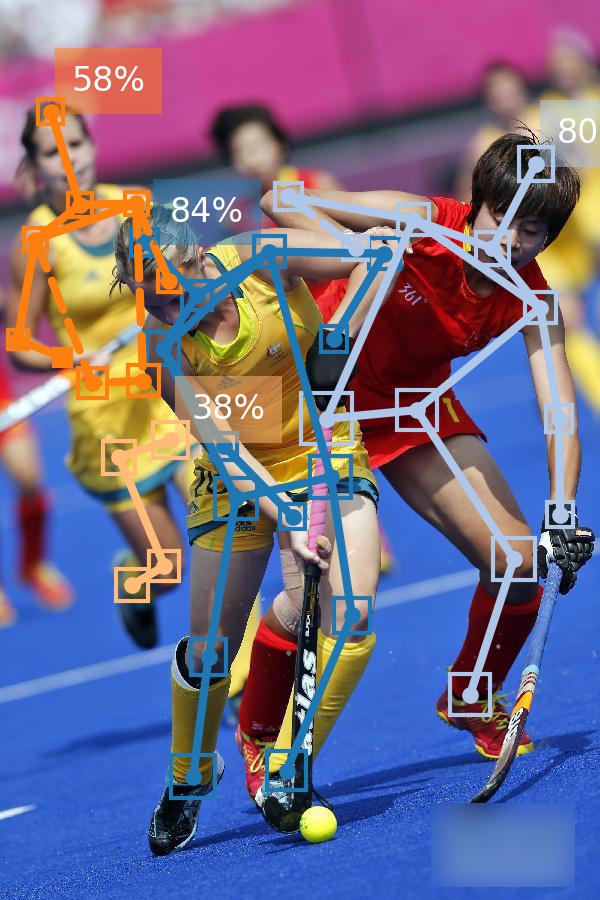}









    \includegraphics[height=5.1cm,trim=0 0 0 0,clip]{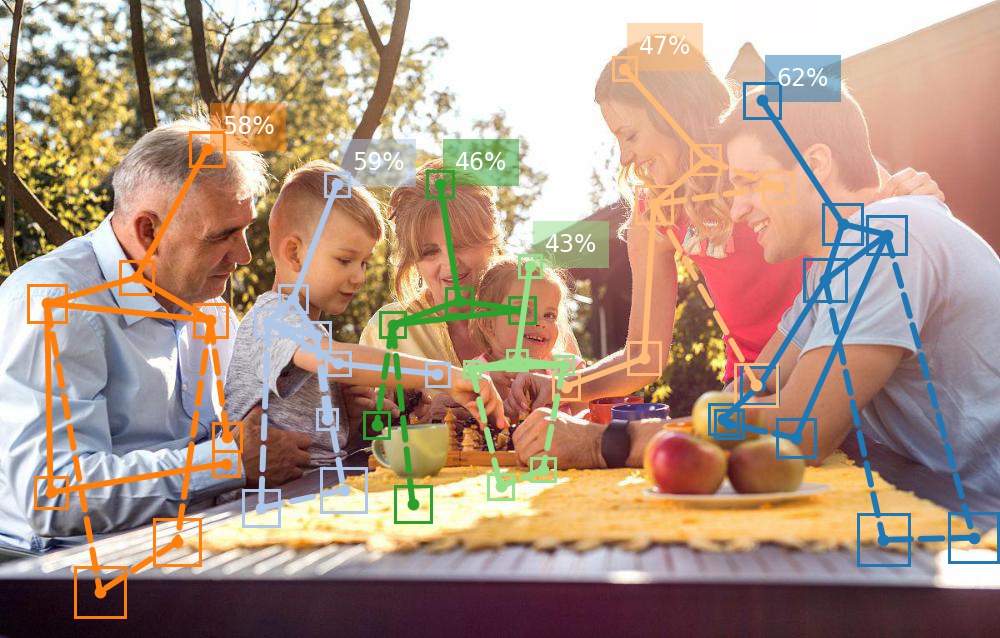}



    \includegraphics[height=5.1cm,trim=0 0 0 0,clip]{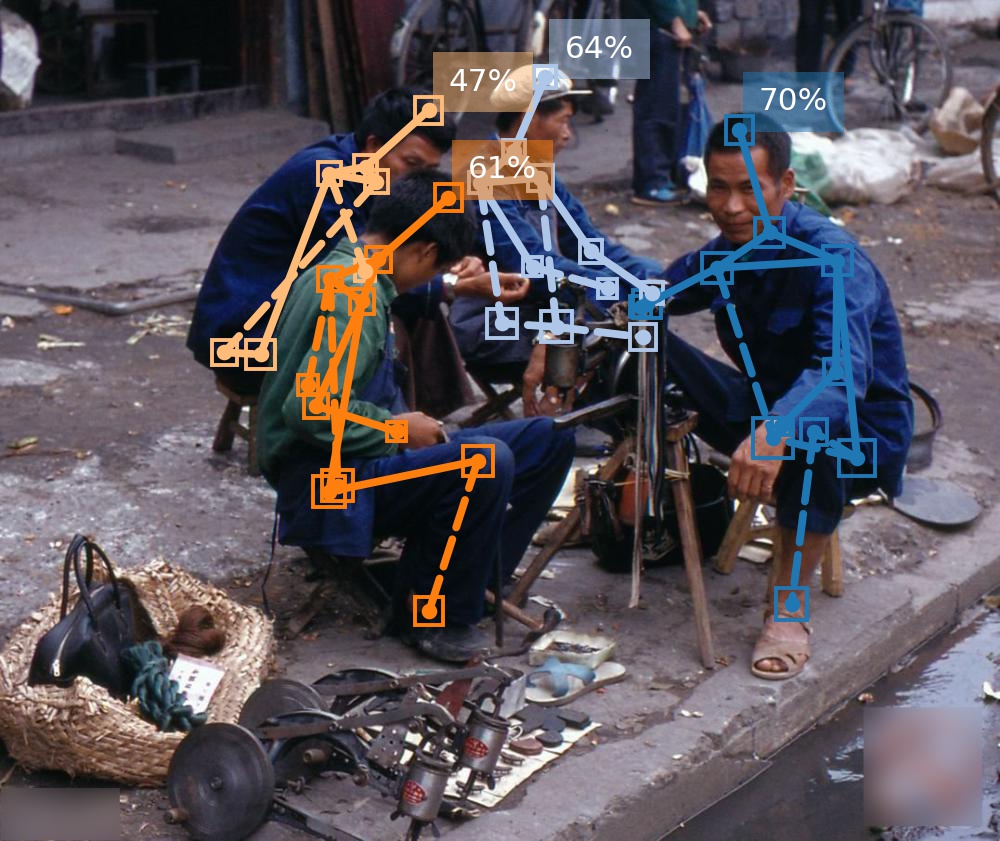}

  \end{tabular}
  \caption{
    Illustration of OpenPifPaf predictions from the
    CrowdPose~\cite{li2019crowdpose} val set with crowd-index \emph{hard} on a
    sports scene, a family photo and a street scene.
  }
  \label{fig:crowdpose-examples}
\end{figure*}

\subsection{Implementation Details}

\paragraph{Neural Network Configuration}
All our models are based on ResNet~\cite{he2016deep} or
ShuffleNetV2~\cite{ma2018shufflenet} base networks and
multiple head networks. The base networks have their input max-pooling
operation removed as it destroys spatial information. The stride from
input image to output feature map is 16 with 2048 features at each
location. We apply no additional modifications to the standard
ResNet models. We use the standard building blocks of
ShuffleNetV2 backbones to construct our custom configurations which we
denote ShuffleNetV2K16/K30. A ShuffleNetV2K16 model has the prediction
accuracy of a ResNet50 with fewer parameters than a ResNet18. The
configuration is specified by the number of output features of the five
stages and the number of repetitions of the blocks in each stage. Our
ShuffleNetV2K16 has output features (block repeats) of 24 (1), 348 (4),
696 (8), 1392 (4), 1392 (1) and our ShuffleNetV2K30 has 32 (1), 512 (8),
1024 (16), 2048 (6), 2048 (1). Spatial $3\times3$ convolutions are
replaced with $5\times5$ convolutions which introduces only a small increase
in the number of parameters because all spatial convolutions are depth-wise.

Each head network is a single $1\times1$ convolution followed
by a sub-pixel convolution~\cite{shi2016real} to double the spatial resolution
bringing the total stride down to eight.
Therefore, the spatial feature map size for an input image of
$801\textrm{px} \times 801\textrm{px}$ is $101\times101$.
The confidence component of a field is normalized with a sigmoid non-linearity
and the scale components for joint-sizes are enforced to be positive with
a softplus~\cite{dugas2000incorporating}.

\paragraph{Augmentations}
We apply the standard augmentations of random horizontal flipping,
random rescaling with a rescaling factor $r \in [0.5, 2.0]$,
random cropping and padding to $385\times385$ followed by color jittering
with 40\% variation in brightness and saturation and 10\% variation in hue.
We also convert a
random 1\% of the images to grayscale and generate strong JPEG compression
artifacts in 10\% of the images.

The tracking task is similarly augmented. The random rescaling is adapted
to an image width in $[0.5\times801, 1.5\times801]$
and random cropping to a maximum image side of 385~px. Half of the image pairs
are randomly reoriented (rotations by multiples of $90^\circ$).
To increase the inter-frame variations, we add a small synthetic camera shift
of maximum 30~px between image pairs.
To further increase the variation, we form image pairs with a random interval of 4, 8 and
12~frames.
In 20\% of image pairs, we replace one of the images with a random image
to provide a higher number of negative samples for tracking.

\paragraph{Single-Image Training}
For ResNet~\cite{he2016deep} backbones, we use
ImageNet~\cite{deng2009imagenet} pretrained models.
ShuffleNetV2~\cite{ma2018shufflenet}
models are trained from random initializations.
We use the SGD~\cite{bottou2010large} optimizer with Nesterov
momentum~\cite{nesterov27method} of 0.95, batch size of 32 and weight decay of $10^{-5}$.
The learning rate is exponentially warmed
up for one epoch from $10^{-3}$ of its target value. At certain
epochs (specified below), the learning
rate is exponentially decayed over 10 epochs by a factor of 10.
We employ model averaging~\cite{polyak1992acceleration,ruppert1988efficient}
to extract stable models for validation. At each optimization
step, we update an exponentially weighted version of the model parameters with a decay
constant of $10^{-2}$.

On CrowdPose, which is a smaller dataset than COCO, we train for 300 epochs.
We set the target learning rate to $10^{-5}$ and
decay at epochs 250 and 280.

On COCO, we use a target learning rate of $10^{-4}$ and decay at epoch 130 and 140.
The training time for 150 epochs of a ShuffleNetV2K16 on two V100 is
approximately 37~hours.
We do not use any additional datasets beyond the COCO keypoint
annotations.

\paragraph{Training for Tracking on PoseTrack}
We use the ShuffleNetV2k30 backbone for all our tracking experiments.
PoseTrack 2018 is a video dataset which means that despite a large number of
annotations, the variation is smaller than in single-image pose datasets.
Therefore, we keep single-image pose estimation on the
COCO dataset~\cite{lin2014microsoft} as an auxiliary task and train on
PoseTrack and COCO simultaneously.
The type of poses that are annotated in the two datasets are similar but not
identical, \textit{e.g.}, one dataset annotates the eyes and the other
does not. During training, we alternate the two tasks between batches.
In one batch we feed pairs of images from the PoseTrack dataset and apply
losses to the corresponding head networks and in the next batch we feed in single
images from COCO and apply losses to the other head networks (see Figure~\ref{fig:model}).
The COCO task is trained identical to the single-image pose estimation
discussed in the previous section, but converted from single images to
pairs of tracked images via synthetic shifts of up to 30px.
Starting from a trained single-image pose backbone,
we train on both datasets with SGD~\cite{bottou2010large} with the same
configuration as for single images. We alternate the dataset every batch
and only do an SGD-step every two batches.
We train for 50 epochs where every epoch consists of 4994 batches.
The training time is 55~minutes per epoch on two V100 GPUs.

\begin{table*}
  \centering
  \caption{Evaluation on the CrowdPose test dataset~\cite{li2019crowdpose}. Our OpenPifPaf result is based on a ResNet50 backbone with single-scale evaluation at 641px. $^{*}$Values extracted from CrowdPose paper~\cite{li2019crowdpose}. $^{+}$Employs multi-scale testing.}
  \label{tab:crowdpose}
  \begin{tabular}{|l|c c c c c c c|}
    \hline
                                              & AP & AP$^{0.50}$ & AP$^{0.75}$ & AP$_\textrm{easy}$ & AP$_\textrm{medium}$ & AP$_\textrm{hard}$ & FPS \\
    \hline\hline
    Mask R-CNN$^{*}$~\cite{he2017mask}        & 57.2 & 83.5    & 60.3   & 69.4 & 57.9 & 45.8   & 2.9 \\
    AlphaPose$^{*}$~\cite{fang2017rmpe}       & 61.0 & 81.3    & 66.0   & 71.2 & 61.4 & 51.1   & 10.9 \\
    HigherHRNet-W48~\cite{cheng2020higherhrnet} & 65.9 & 86.4  & 70.6   & 73.3 & 66.5 & 57.9   & - \\
    SPPE~\cite{li2019crowdpose}               & 66.0 & 84.2    & 71.5   & 75.5 & 66.3 & 57.4   & 10.1 \\
    HigherHRNet-W48$^{+}$~\cite{cheng2020higherhrnet} & 67.6 & 87.4 & 72.6  & 75.8 & 68.1 & 58.9  & - \\

    \textbf{OpenPifPaf} (ours)
    & \textbf{70.5} & \textbf{89.1} & \textbf{76.1}  & \textbf{78.4} & \textbf{72.1} & \textbf{63.8}   & \textbf{13.7} \\
    \hline
  \end{tabular}
\end{table*}

\subsection{Results}

\paragraph{Crowded Single-Image Pose Estimation}
In Figure~\ref{fig:crowdpose-examples}, we show example pose predictions from
the CrowdPose~\cite{li2019crowdpose} validation set. We show results
in a diverse selection of sports disciplines and everyday settings. All shown
images are from the \emph{hard} subset with a crowd-index larger than 0.8.

In Table~\ref{tab:crowdpose}, we show a quantitative comparison of our
performance with other methods. We are not only more precise across all
precision metrics AP, AP$^{0.50}$, AP$^{0.75}$, AP$_\textrm{easy}$,
AP$_\textrm{medium}$ and AP$_\textrm{hard}$ but also predict faster than all
previous top-performing methods at 13.7 FPS (frames-per-second)
on a single GTX1080Ti.

\begin{table}[t]
  \centering
  \caption{
    Evaluation metrics for the COCO 2017 test-dev dataset for
    bottom-up methods. Numbers are extracted from the respective papers.
    Our prediction time is determined on a single V100 GPU.
    $^*$Only evaluating images with three person instances.
  }
  \label{tab:high-res}
  \begin{tabular}{|l|c c c c|}
    \hline
                            & \textbf{AP} & AP$^{M}$ & AP$^{L}$ & $t$ [ms] \\
    \hline\hline
    OpenPose~\cite{cao2017realtime}                 & 61.8 & 57.1 & 68.2 & 100 \\
    Assoc. Emb.~\cite{newell2017associative}  & 65.5 & 60.6 & 72.6 & 166 \\
    PersonLab~\cite{papandreou2018personlab}     & 68.7 & 64.1 & 75.5 & - \\
    MultiPoseNet~\cite{kocabas2018multiposenet}  & 69.6 & 65.0 & 76.3 & 43$^*$ \\
    HigherHRNet~\cite{cheng2020higherhrnet}           & 70.5 & 66.6 & 75.8 & $>$1000 \\
    \textbf{OpenPifPaf} (ours)  & \textbf{\hl{71.9}} & \textbf{\hl{68.5}} & \textbf{\hl{77.4}} & \hl{69} \\
    \hline
  \end{tabular}
\end{table}

\paragraph{COCO}
All state-of-the-art methods compare their performance on the well-established
COCO keypoint task~\cite{lin2014microsoft}.
Our quantitative results on the private 2017 test-dev set are shown
in Table~\ref{tab:high-res} along with other bottom-up methods.
This comparison includes field-based
methods~\cite{cao2017realtime,papandreou2018personlab,kreiss2019pifpaf}
and methods based on associative
embedding~\cite{newell2017associative,cheng2020higherhrnet}.
We perform on par with the best existing
bottom-up method.
We evaluate on rescaled images where the longer edge is 801~px which is the
same image size that will be used for tracking below. We evaluate a single
forward pass without horizontal flipping and without multi-scale evaluation because
we aim for a fast method.
The average time per image with a GTX1080Ti is 152~ms (63~ms on a V100) of which 29~ms
is used for decoding.

\paragraph{Pose Tracking}

\begin{table}
  \centering
  \caption{
    Evaluation metrics on the test sets of
    (a) PoseTrack 2018~\cite{andriluka2018posetrack} and
    (b) PoseTrack 2017~\cite{iqbal2017posetrack}.
    Numbers are extracted from the respective papers
    and the leaderboard.
    All methods are online methods apart from DetTrack~\cite{wang2020combining}.
  }
  \label{tab:posetrack}
  \subfloat[]{
    \begin{tabular}{|l|c c|H H c|}
      \hline
        PoseTrack 2018 & \textbf{MOTA} & FPS & wrists AP & ankles AP & AP \\
      \hline\hline
      openSVAI~\cite{ning2018top}      & 54.5 & - & 59.2 & 56.7 & 63.1 \\
      MIPAL~\cite{hwang2019pose}       & 54.9 & - & 60.2 & 56.9 & 67.8 \\
      Miracle~\cite{yu2018multi}       & 57.4 & - & 68.2 & 66.1 & 70.9 \\
      MSRA/FlowTrack~\cite{xiao2018simple}       & 61.4 & 0.7 & 73.0 & 69.1 & 74.0 \\
      \textbf{OpenPifPaf} (ours)        & \textbf{61.7} & \textbf{12.2} & - & - & 71.9 \\
      \hline
    \end{tabular}
  }

  \subfloat[]{
    \begin{tabular}{|l|c c|H H c|}
      \hline
        PoseTrack 2017 & \textbf{MOTA} & FPS & wrists AP & ankles AP & AP \\
      \hline\hline
      STAF~\cite{raaj2019efficient}        & 53.8 & 3 & 65.02 & 60.72 & 70.3 \\
      MIPAL~\cite{hwang2019pose}           & 54.5 & - & 60.94 & 56.04 & 68.8 \\
      MSRA/FlowTrack~\cite{xiao2018simple}      & 57.8 & 0.7 & 71.52 & 65.7 & 74.6 \\
      HRNet~\cite{sun2019deep}             & 57.9 & - & 72.04 & 66.96 & 74.9 \\
      LightTrack~\cite{ning2020lighttrack} & 58.0 & - & 64.58 & 58.43 & 66.6 \\
      \textbf{OpenPifPaf} (ours)        & 60.6 & \textbf{12.2} & - & - & 71.5 \\
      KeyTrack~\cite{snower201915}     & \textbf{61.2} & 1.0 & 71.91 & 64.95 & 74.0 \\
      \hline
      DetTrack (offline)~\cite{wang2020combining} & 64.1 & - & - & - & 74.1 \\
      \hline
    \end{tabular}
  }
\end{table}

We want to track multiple human poses in videos. We train and validate
on the PoseTrack~2018 dataset~\cite{andriluka2018posetrack}.
Table~\ref{tab:posetrack} shows our main results for pose tracking
on both of the private test sets of Posetrack 2017 and 2018.
\hl{We also show our single-image average precision (AP) which highlights
that our performant tracking method can compensate for a lower AP, \eg,
compared to MSRA/FlowTrack~\cite{xiao2018simple}, and still outperform
in overall MOTA and FPS.}
All our results are produced in a single pass and online (without future frames).
The frames per second~(FPS) stated in
Table~\ref{tab:posetrack} refer to the single process, sequential
evaluation.
In addition, we provide extra metrics that are not published
on the leaderboards. For PoseTrack~2017, our MOTP is 84.5, precision is 84.1
and recall is 77.7. For PoseTrack~2018, our MOTP is 84.9, precision is 84.4
and recall is 78.3.

\begin{figure}
  \centering
    \includegraphics[height=3.2cm,trim=0cm 0.5cm 0cm 0.5cm,clip]{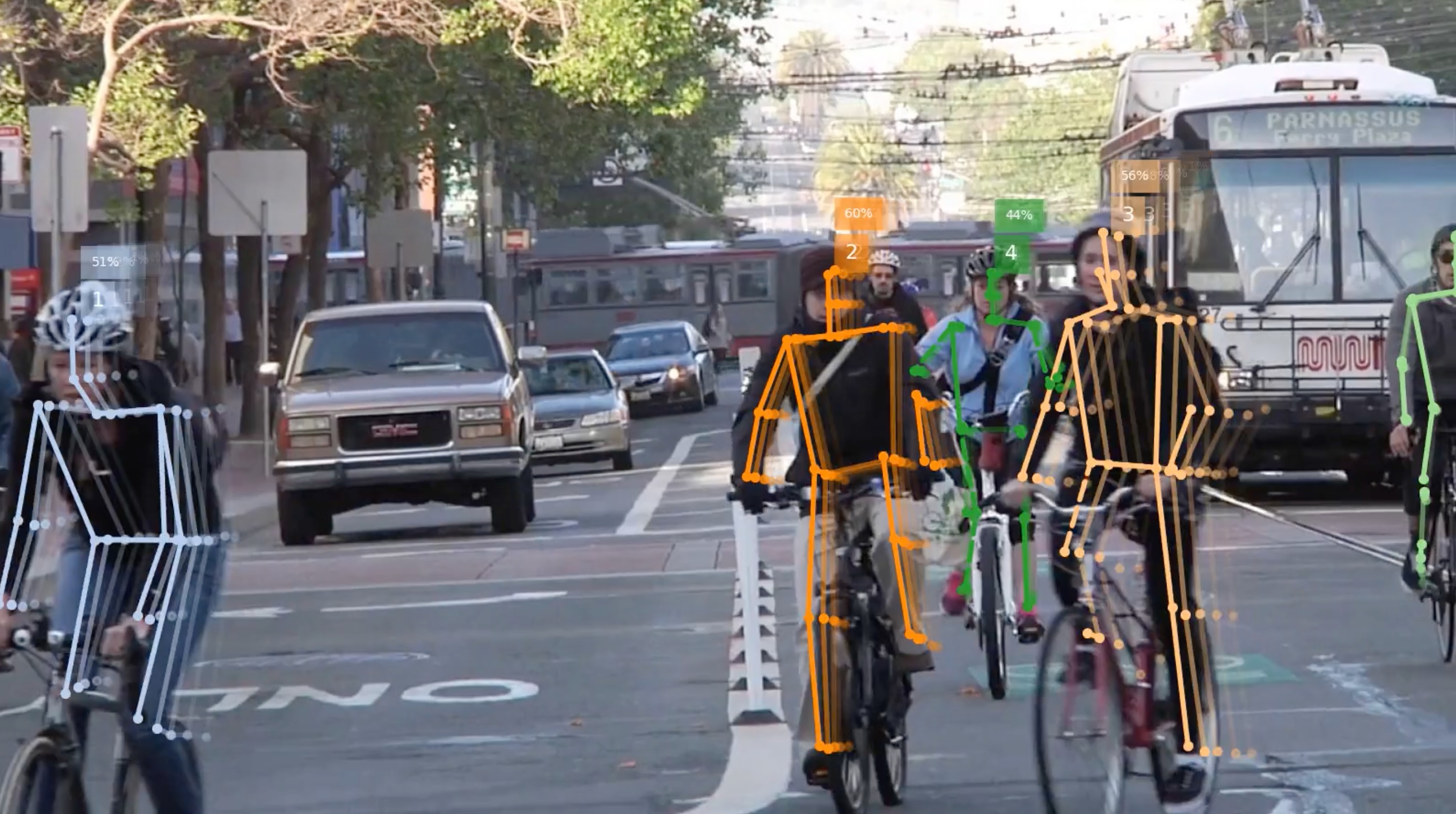}
    \includegraphics[height=3.2cm,trim=0cm 0cm 0 0cm,clip]{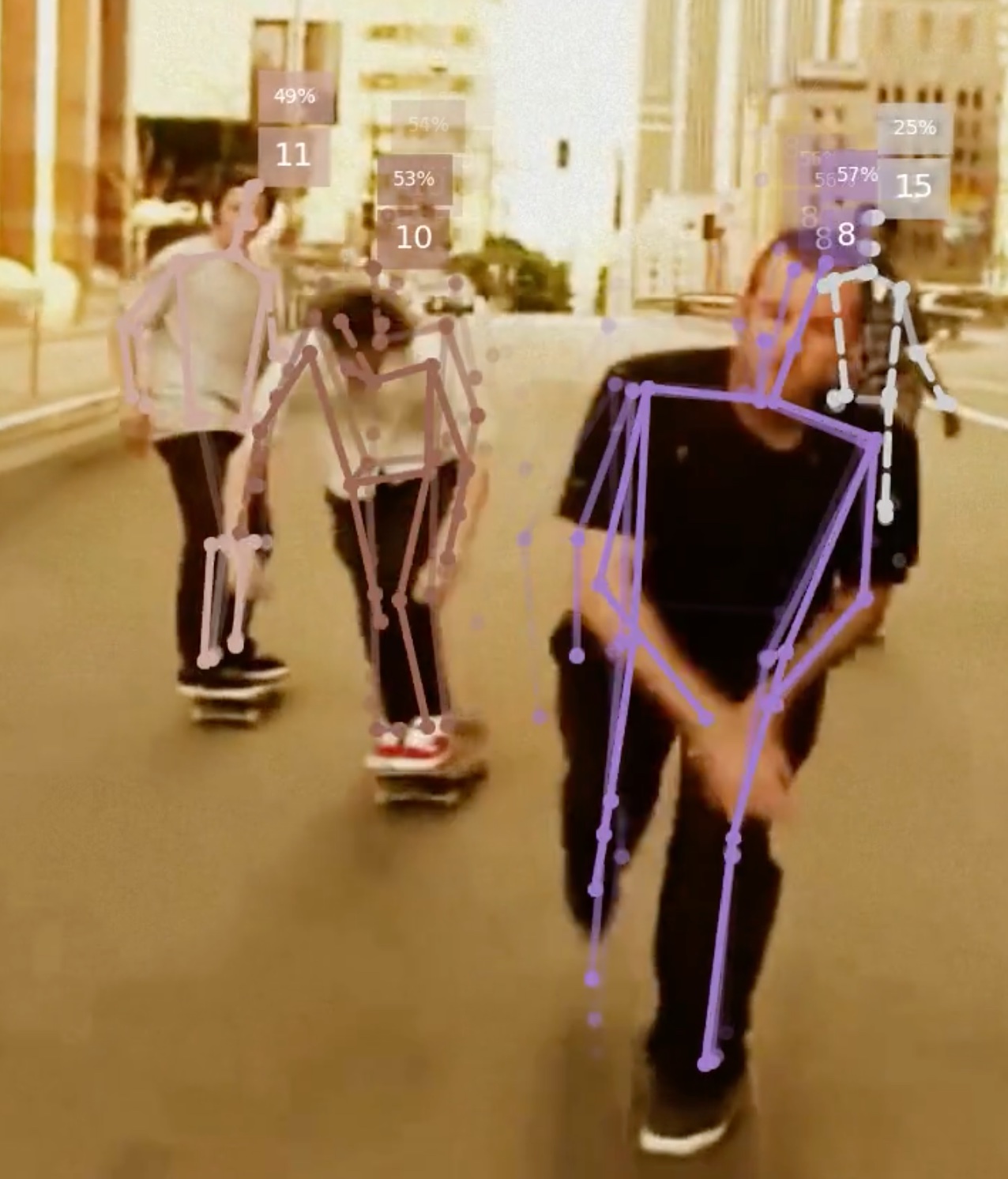}
  \caption{
    Qualitative results from the Posetrack 2018~\cite{andriluka2018posetrack}
    validation set.
    Images show tracks of spatio-temporal poses including their frame-to-frame
    associations where only connections that
    were used to construct the poses are shown.
  }
  \label{fig:qualitative}
\end{figure}

\begin{figure}
  \centering
    \includegraphics[height=5.0cm,trim=0 0 0 0,clip]{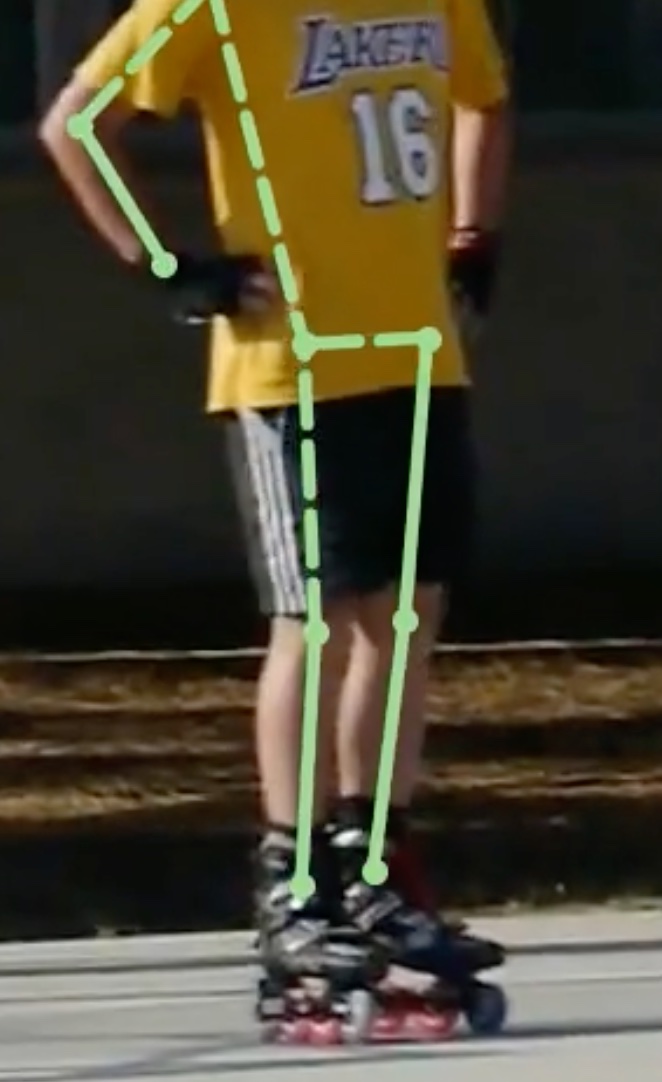}
    \includegraphics[height=5.0cm,trim=0 0 0 0,clip]{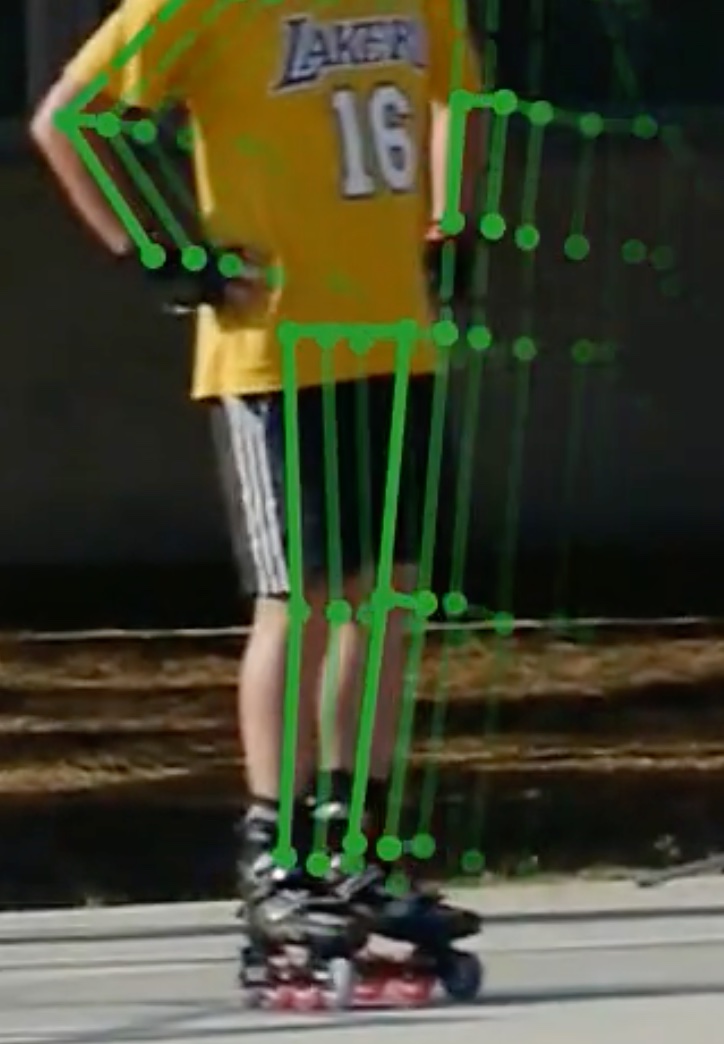}
  \caption{
    Qualitative results from the Posetrack 2018~\cite{andriluka2018posetrack}
    validation set.
    Left:~Single-image detection. The person's left shoulder is not visible and
    therefore the left arm cannot be connected to the rest of the body.
    Right:~Spatio-temporal pose. Multiple temporal connections allow to safely
    connect both left and right arm to the rest of the body.
  }
  \label{fig:qualitative-hungarian-vs-tcaf}
\end{figure}

\begin{figure}
  \centering
    \includegraphics[width=\linewidth,trim=0cm 0cm 0cm 0cm,clip]{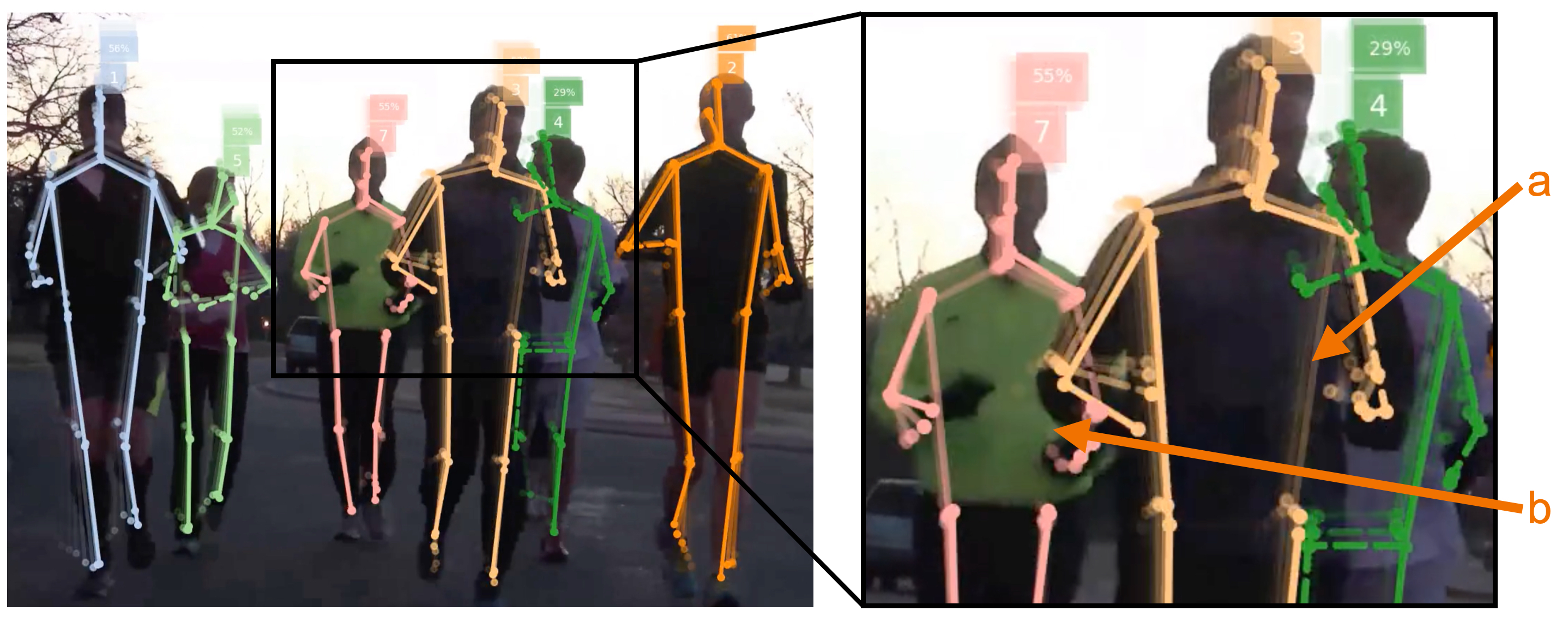}
  \caption{
    Qualitative results from the Posetrack 2018~\cite{andriluka2018posetrack}
    validation set.
    (a)~indicates a connection that has been made spatially in a previous frame
    but for the last few frames the left leg of person 3 is connected to the rest
    of the body only through temporal connections.
    (b)~shows a connection that is temporarily occluded by the arm of the person
    in front and also here our algorithm decided to connect the left leg via
    temporal connections instead of spatial ones.
  }
  \label{fig:qualitative-spatiotemp}
\end{figure}

Spatio-temporal poses on real-world examples are shown in
Figure~\ref{fig:qualitative}. They show challenging scenarios
with occlusions.
Figure~\ref{fig:qualitative-hungarian-vs-tcaf} highlights the ability of spatio-temporal poses to complete
poses through time, \ie, even when a pose is partitioned because of occlusion
in the current frame, multiple temporal connections (TCAF) form a single tracked pose.
Similarly, for the poses 3, 4, 5 and 7 in Figure~\ref{fig:qualitative-spatiotemp},
the
associations from shoulders to hips are often difficult because of the lighting
condition.
Depending on the predicted association confidences, the decoder determines
automatically whether to connect to a keypoint with a spatial or temporal
connection.
In these difficult scenarios, the greedy decoder
completed these poses with multiple temporal connections~(TCAF).

\begin{figure*}
  \centering
    \includegraphics[width=0.45\linewidth]{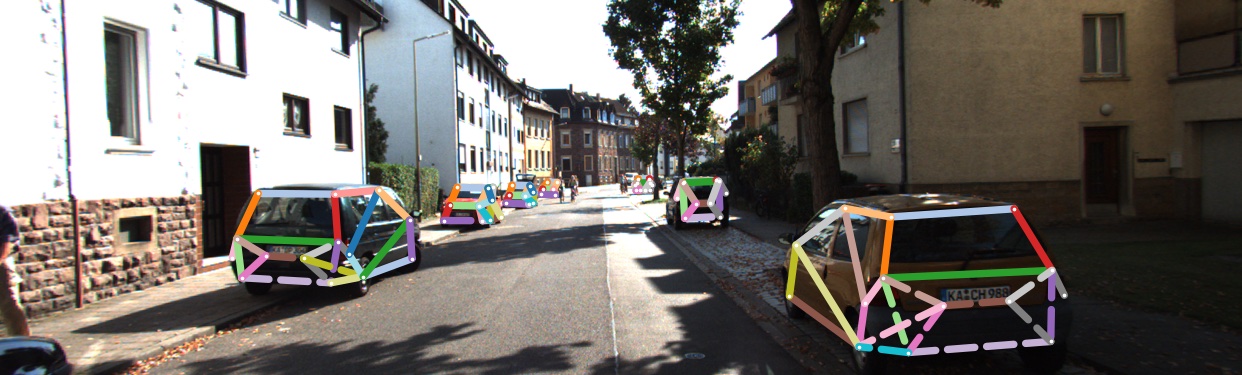}
    \includegraphics[width=0.45\linewidth,trim=0 10cm 0 49.5cm,clip]{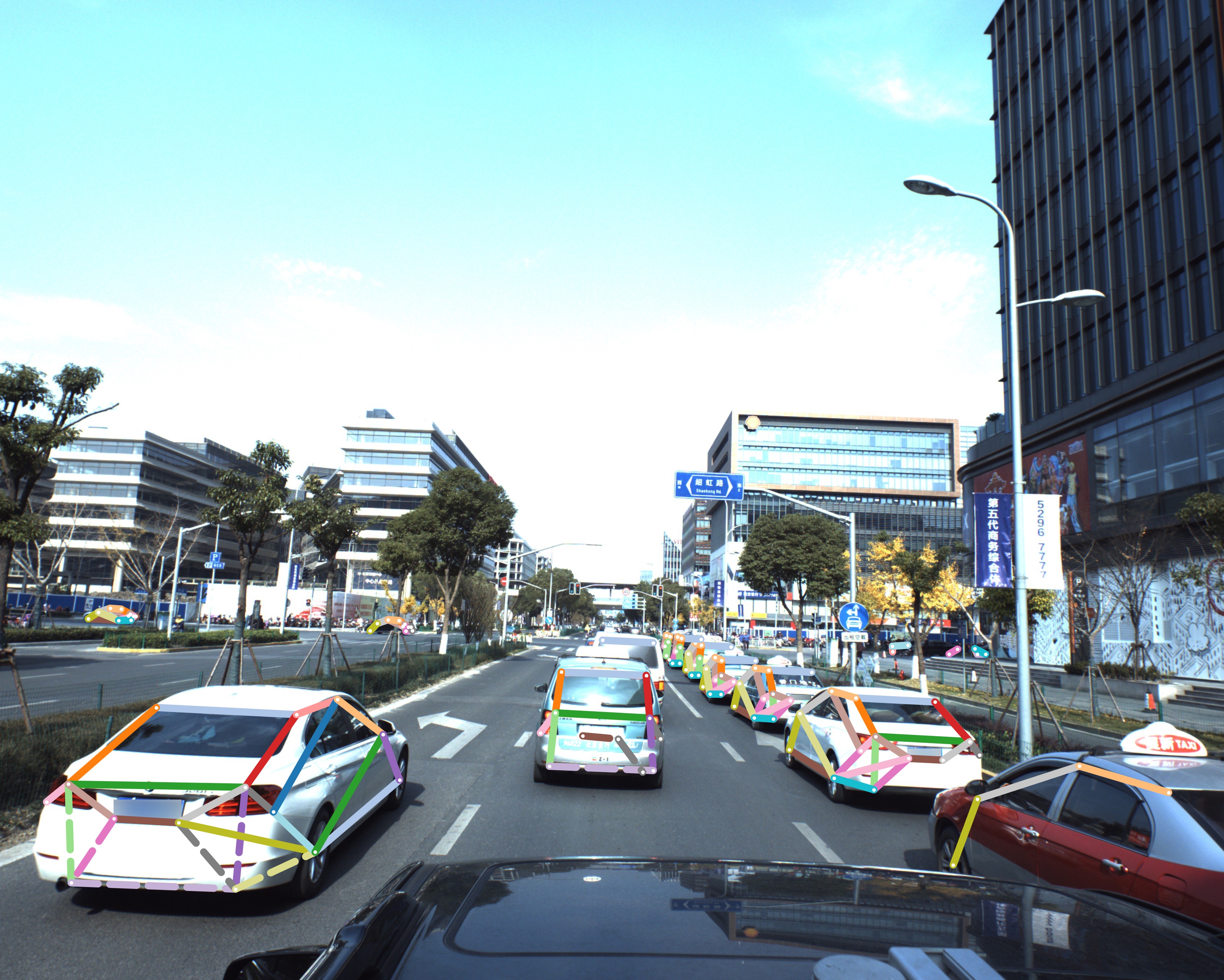}
    \includegraphics[width=0.45\linewidth]{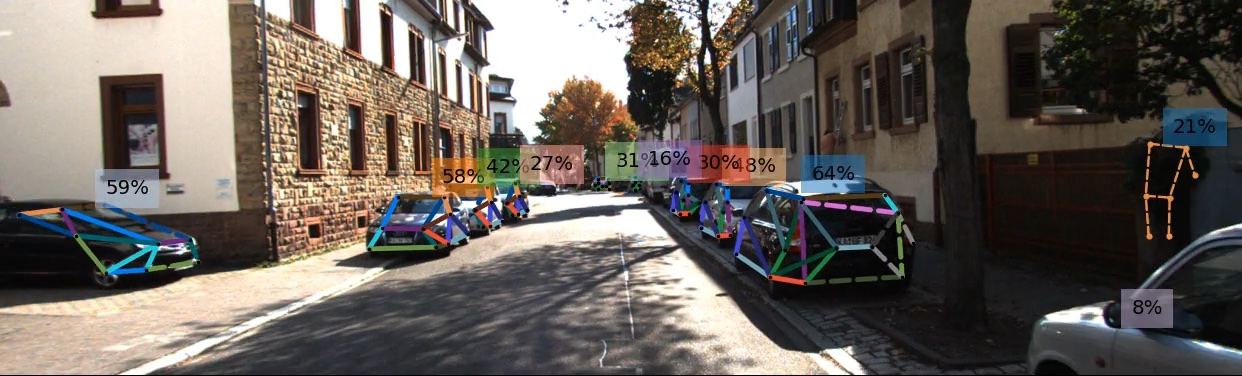}
    \includegraphics[width=0.45\linewidth]{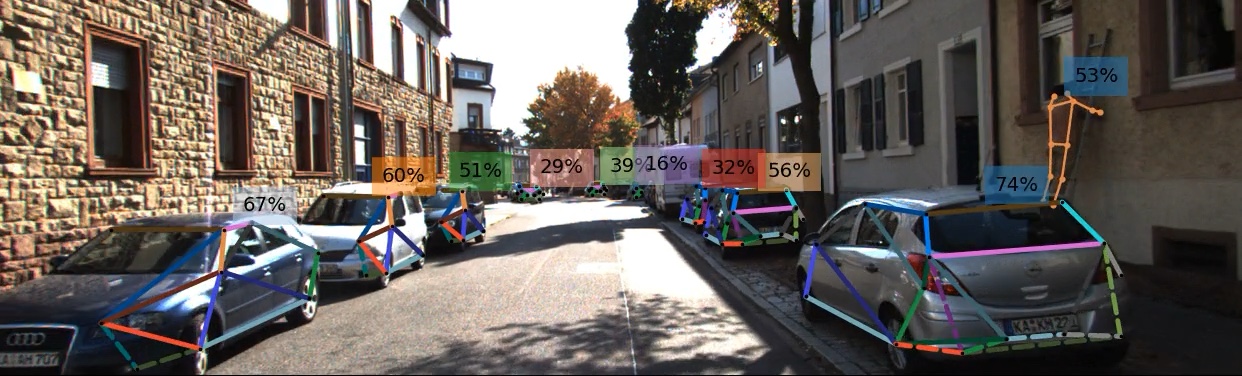}
    \includegraphics[width=0.45\linewidth]{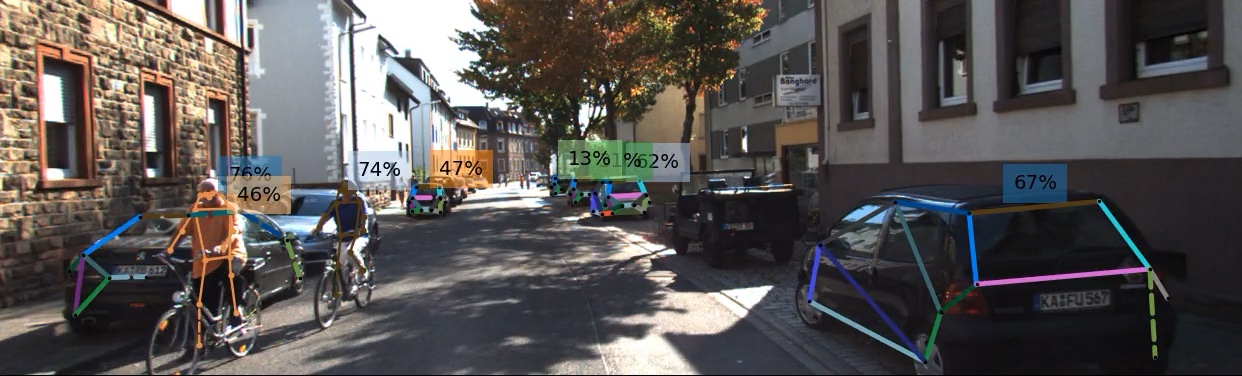}
    \includegraphics[width=0.45\linewidth]{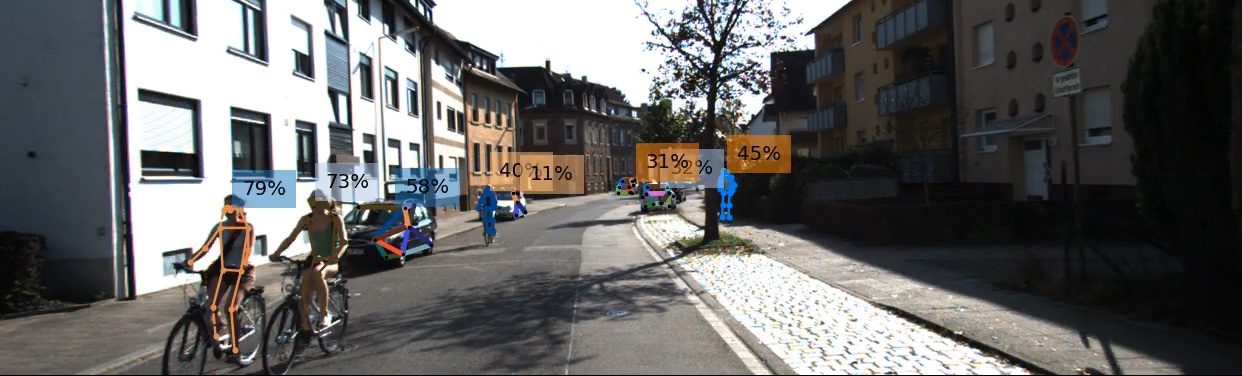}
    \includegraphics[width=0.45\linewidth]{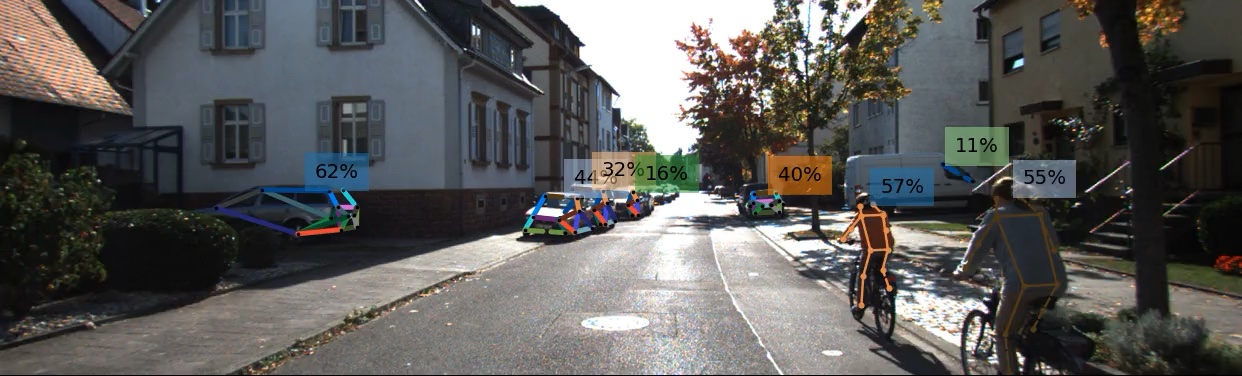}
    \includegraphics[width=0.45\linewidth]{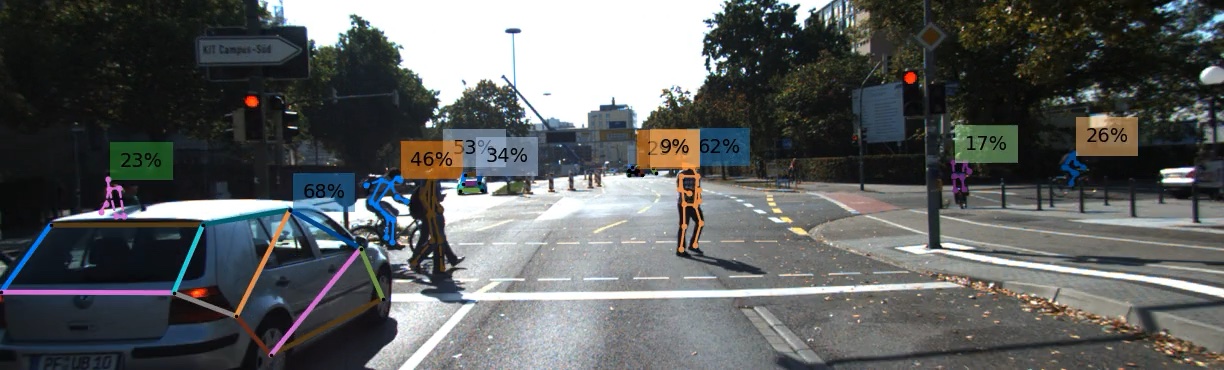}
    \includegraphics[width=0.45\linewidth]{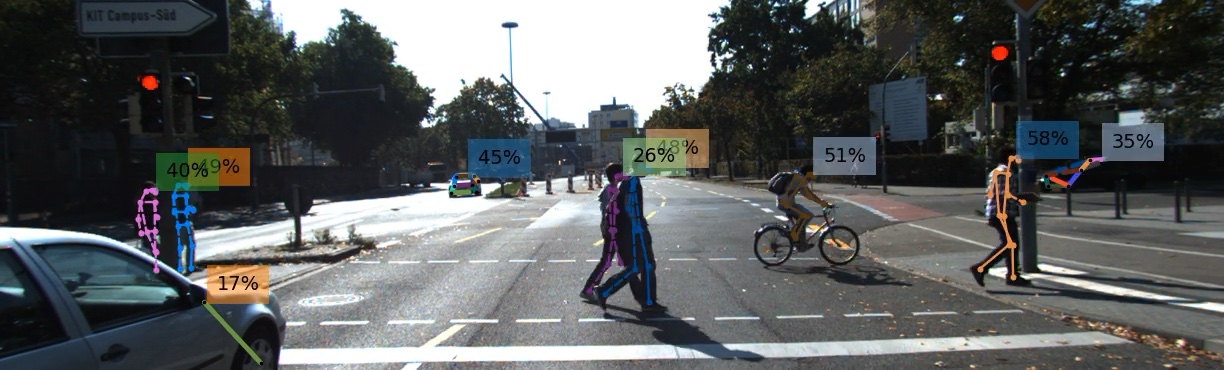}
    \includegraphics[width=0.45\linewidth]{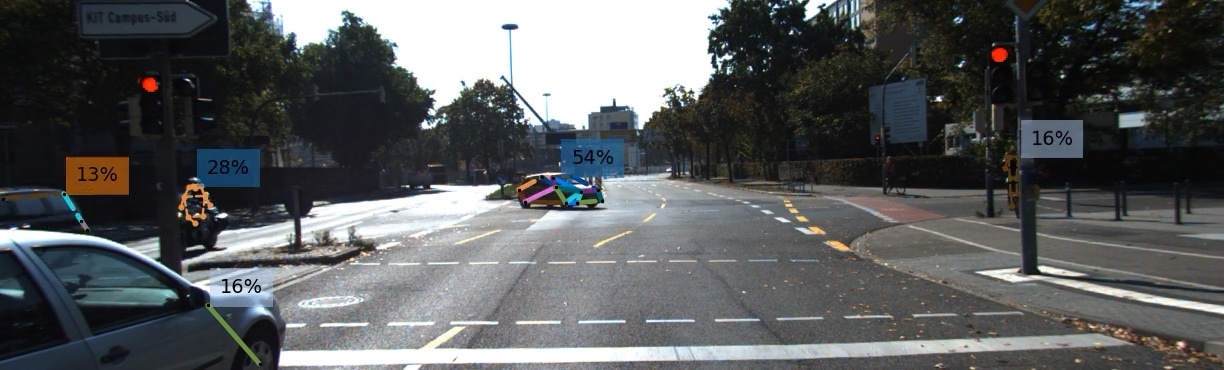}
  \caption{
    Qualitative results from the KITTI \cite{kitti} and ApolloCar3D \cite{song2019apollocar3d} datasets. We resolve distant pedestrians, cyclists and cars and handle changing lighting conditions well.
  }
  \label{fig:cars}
\end{figure*}

\begin{figure}
  \centering
    \includegraphics[height=3.5cm,trim=0 0 0 0cm,clip]{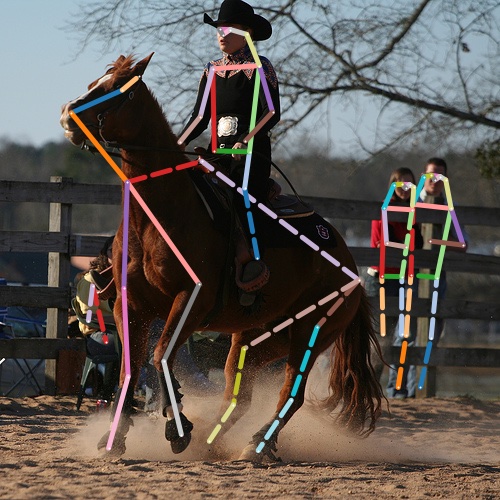}
    \includegraphics[height=3.5cm,trim=0 0cm 0 0cm,clip]{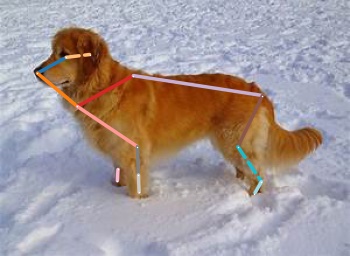}
  \caption{
    Qualitative results from the Animal-Pose dataset \cite{cao2019cross}. The left image was processed by a person model and an animal model.
  }
  \label{fig:animals}
\end{figure}

\begin{table}
  \centering
  \caption{Quantifying detection performance for pedestrians, cars and animals. In the ``Pedestrians'' column, we show the detection rate on KITTI~\cite{kitti} with IoU=0.3 and instance threshold of 0.2 for all methods.
  For ``Vehicles'', we show the keypoint detection rate on ApolloCar3D~\cite{song2019apollocar3d} which was published in previous methods and we also provide AP in the text.
  In the ``Animals'' column, we provide keypoint AP as defined in the Animal-Pose dataset~\cite{cao2019cross}.}
  \label{tab:ped-veh-ani}
  \begin{tabular}{|l|c c c|}
    \hline
    Method                                    & Pedestrians & Vehicles & Animals \\
    \hline
    Mono3D~\cite{chen2017multi}               & 73.2 & - & - \\
    3DOP (stereo)~\cite{chen20153d}           & 73.1 & - & - \\
    MonoDIS~\cite{simonelli2019disentangling} & 60.5 & - & - \\
    SMOKE~\cite{liu2020smoke}                 & 39.1 & - & - \\
    MonoPSR~\cite{ku2019monocular}            & 82.8 & - & - \\
    CPM~\cite{wei2016convolutional}           & - & 75.4 & - \\
    WS-CDA~\cite{cao2019cross}                & - & - & 44.3 \\
    \textbf{OpenPifPaf} (ours)  & \textbf{84.6} & \textbf{86.1} & \textbf{47.8} \\
    \hline
    Human labelers~\cite{song2019apollocar3d} & - & 92.4 & - \\
    \hline
  \end{tabular}
\end{table}

\paragraph{Pedestrian, Car and Animal Poses}
A holistic perception framework for autonomous vehicles also needs to be
able to generalize to other classes than humans. We show that we can predict
poses of cars and animals with high accuracy in Figures~\ref{fig:cars}
and~\ref{fig:animals} and provide a quantitative summary in Table~\ref{tab:ped-veh-ani}.

On car instances, our model achieves an average precision~(AP) of 76.1\%.
The AP metric follows the same protocol of human instances,  but to the best
of our knowledge no previous method has evaluated AP on
ApolloCar3D~\cite{song2019apollocar3d} without leveraging 3D information.
Hence, we include a study on the keypoint detection rate, which has been defined
in the ApolloCar3D dataset~\cite{song2019apollocar3d} and considers a keypoint to
be correctly estimated if the error is less than 10 pixels. Our method achieves
a detection rate of 86.1\% compared to 75.4\% of CPM~\cite{wei2016convolutional}.
Notably, the authors of ApolloCar3D~\cite{song2019apollocar3d} also report the
detection rate of the human labelers to be 92.4\%.

On animal instances, our model achieves an AP of 47.8\%, compared to 44.3\% of
WS-CDA, the baseline developed by the authors of the Animal-Pose
dataset~\cite{cao2019cross}.  Lower performances on animals are due to the
smaller dataset size with just 4K training instances. Simultaneous training for
humans and animals to achieve better generalization is left for future work.

\subsection{Ablation Studies}

\begin{table*}
  \centering
  \caption{
    Ablation studies of skeleton choice and decoder configurations for single-image pose estimation.
    All results (except where explicitly stated otherwise) are produced with the same ShuffleNetV2k16
    model on the COCO val set~\cite{lin2014microsoft} on a single GTX1080Ti.
    First, we review different backbone architectures (a ResNet50~\cite{he2016deep}
    and a larger ShuffleNetV2~\cite{ma2018shufflenet}).
    Second, we show that only using confident keypoints leads to a large
    drop in precision.
    Third, we observe that the Frontier decoder is more important for denser skeletons while incurring almost no overhead on sparse skeletons.
    Fourth, we can produce a memory-efficient version of our decoder at a cost of 1.4\% in AP. The biggest drop in accuracy comes from not rescoring the CAF field and the largest contributor to increasing the inference time is not rescoring the seeds.
  }
  \label{tab:single-image-ablation}
  \begin{tabular}{|l l|H H c c c c c c c|}
    \hline
    & & CIF-NMS & HR & AP & AP$^{0.50}$ & AP$^{0.75}$ & AP$^M$ & AP$^L$ & $t$ [ms] & $t_\textrm{dec}$ [ms] \\
    \hline\hline

    & \textbf{original} (ShuffleNetV2K16)
    & - & $\checkmark$ & \textbf{66.8} & \textbf{86.5} & \textbf{73.2} & \textbf{62.1} & \textbf{74.6} & \textbf{50} & \textbf{19} \\

    \hline

    Backbone & ResNet50
    & - & $\checkmark$      & 68.2 & 87.9 & 74.6 & 65.8     & 72.7     & 64   & 22 \\
    & ShuffleNetV2K30
    & - & $\checkmark$      & 71.0 & 88.8 & 77.7 & 66.6     & 78.5     & 92   & 16 \\

    \hline

    Keypoints & independent-only
    & - & $\checkmark$ & -8.1 & -6.3 & -9.5 & -8.7 & -7.3 & $\pm0$ & $\pm0$ \\

    \hline

    Frontier decoder & no-frontier
    & - & $\checkmark$ & $\pm0.0$ & -0.1 & +0.1 & $\pm0.0$ & -0.1 & -1 & -1 \\
    & dense
    & - & $\checkmark$ & +0.1 & +0.2 & +0.2 & -0.3 & +0.5 & +15 & +15 \\
    & no-frontier and dense
    & - & $\checkmark$ & -0.3 & +0.1 & -0.1 & -0.5 & $\pm0.0$ & +14 & +14 \\

    \hline

    memory efficient & no seed rescoring
    & - & $\checkmark$ & -0.1 & -0.4 & -0.1 & +0.2 & +0.1 & +71 & +54 \\
    & no seed rescoring (with NMS)
    & $\checkmark$ & $\checkmark$ & +0.1 & +0.1 & $\pm0.0$ & +0.2 & +0.0 & +19 & +15 \\
    & no CAF rescoring
    & - & $\checkmark$ & -1.0 & -0.3 & -1.0 & -1.0 & -1.7 & -1 & -1 \\
    & no rescoring (with NMS), without HR
    & $\checkmark$ & - & -1.4 & -0.4 & -1.4 & -1.0 & -2.3 & +9 & +7 \\

    \hline
  \end{tabular}
\end{table*}

We study the impact of the backbone, the precise criteria for a keypoint,
our proposed Frontier decoder, a memory efficient
decoder, alternatives to TCAF and the impact of input image size.
We start with studies for single images on the COCO val set
(Table~\ref{tab:single-image-ablation}) before moving to
tracking studies for PoseTrack (Table~\ref{tab:posetrack-result2018val}).

Our single-image studies are run with an option to force complete poses. This is the
common practice as the COCO metric does not penalize false positive
keypoints within poses. This option would not be used in most real-world
settings. Without forcing complete poses, the decoding time and
the total prediction time is reduced by about 10ms.

\paragraph{Backbone}
The reference backbone is a small ShuffleNetV2K16. We show comparisons to the
larger ResNet50 and ShuffleNetV2K30 backbones and show how they improve
precision (AP) and at what cost in timing.

\paragraph{Keypoint Criterium}
We try to illuminate why our precision and speed is significantly better than
OpenPose~\cite{cao2017realtime}. OpenPose first detects keypoints and then
associates them. Therefore, every keypoint has to be detectable individually.
In OpenPifPaf, new keypoint associations are generated from a source keypoint.
These new keypoints are not previously known. They are discovered in the
association. That allows OpenPifPaf to generate poses from a strong seed
keypoint and connect to less confident keypoints. In ``independent-only'',
we restrict the keypoints of OpenPifPaf to be all of the quality of
an independent seed keypoint and observe a dramatic drop of 8.1\% in AP.

\paragraph{Frontier Decoder}
Next, we study the impact
of the Frontier decoder with respect to a simpler decoder without frontier.
The standard pose is sparsely connected and, therefore, the frontier only
has few alternatives to prioritize. For a denser pose (``dense''), the
impact of the frontier (compare with ``no-frontier and dense'') is more
pronounced (+0.3~AP).

\paragraph{Memory Efficient Decoding}
In the bottom part of Table~\ref{tab:single-image-ablation}, we study
the effect of removing the high-resolution accumulation map (HR)
to reduce the memory footprint.
This high resolution map is used in two places. First, to rescore the seeds
and, second, to rescore the CAF. The impact of the seed rescoring
is only 0.1 in AP but comes at a large cost in decoding time.
As an alternative, we investigate a local non-maximum suppression (NMS)
that selects a seed only if it is the highest confidence in a $3\times3$
window (introduced in CenterNet~\cite{zhou2019objects}). This NMS reduces
the decoding time but not back to the original speed.
Independently, we study the
impact of rescoring the CAF field which is about +1.0\% in AP. Only
when both the seed rescoring and the CAF rescoring are removed, the
creation of the HR maps can be omitted. In that memory efficient
configuration (bottom line in Table~\ref{tab:single-image-ablation}),
the AP dropped by 1.4\% with respect to ``original''. This demonstrates
the importance of the high-resolution accumulation for speed and accuracy
and which should only be removed when absolutely necessary.

\begin{table}
  \centering
  \caption{
    Baselines and ablation studies on the PoseTrack 2018 validation set~\cite{andriluka2018posetrack} on a single V100 GPU.
    We outperform Hungarian trackers with euclidean and
    OKS distance functions in accuracy for a small overhead in FPS.
    We also study our sensitivity to the input image size. For image sizes of 513px, we observe a drop of 2.9 in MOTA
    but run 82\% faster at 22.2 FPS.
  }
  \label{tab:posetrack-result2018val}
  \begin{tabular}{|l l|c H H H c|}
    \hline
       & & \textbf{MOTA} & wrists AP & ankles AP & total AP & FPS \\
    \hline\hline
    & \textbf{original} (801px)    & \textbf{66.4} & 66.2     & 65.4     & 74.7     & 12.2 \\

    \hline

    Hungarian
    & euclidean  & -1.5 & -1.9 & -1.3 & -1.1 & +4\% \\
    & OKS        & -2.0 & -    & -   & -   & +1\% \\

    \hline

    Image size
    & 513px          & -2.9 & -   & -   & -   & \textbf{+82\%} \\
    & 641px          & -0.9 & -   & -   & -   & +37\% \\
    & 1201px         & -1.7 & -   & -   & -   & -49\% \\

    \hline
  \end{tabular}
\end{table}

\paragraph{Tracking Baselines}

We conducted detailed studies of our method on the Posetrack 2018 validation set that are
shown in Table~\ref{tab:posetrack-result2018val}.
First, we created two baselines ourselves. Both baselines first do single-image
pose estimation and then use the Hungarian algorithm~\cite{kuhn1955hungarian} to track poses
from frame to frame. Our first algorithm uses a simple Euclidean distance
between joints to construct a pose similarity score. Our second method
replaces the Euclidean distance with an OKS-based distance that is used in the COCO metric to compare predictions to ground truth. Both methods show a drop in MOTA of 1.5 and 2.0 while operating at about the same speed as our ``original'' model.
This demonstrates that the overhead of our tracking network
is comparable to the small overhead of the Hungarian algorithm with
respect to the single-image model.

\paragraph{Tracking Ablation}

We studied the effect of input image size at the bottom
of Table~\ref{tab:posetrack-result2018val}.
Our ``original'' model rescales the image width to 801px.
Larger images do not show an improvement in accuracy (MOTA) while
becoming significantly slower. Smaller input images
decrease MOTA but at the same time can drastically increase
speed. Most applications can probably tolerate an accuracy
reduction by 0.9 in MOTA to improve speed by +37\%. When the
input image size is reduced to 513px, MOTA drops by 2.9 (still a
great result) which comes with a speed improvement of +82\% to
a fast 22.2~FPS.

\section{Conclusions}

We have demonstrated a new method for bottom-up pose tracking for 2D human poses
and shown its strength in crowded and occluded scenes that are relevant for
perception in self-driving cars and social robots. We outperform previous
state-of-the-art methods on CrowdPose and on PoseTrack2018. On PoseTrack2017
we are on par with the state-of-the-art but run an order of magnitude faster.
We have also
shown that our method generalizes to pose estimation of cars and animals.
We can run all versions simultaneously on an image sequence and form the union of the predictions. In the future, we can investigate shared backbone
architectures
to create a holistic perception framework for autonomous vehicles.

\section{Acknowledgements}

This work was supported by the Swiss National Science Foundation under the Grant 2OOO21-L92326 and the SNSF Spark fund (190677). We also thank our lab members and reviewers for their valuable comments.


%




\ifCLASSOPTIONcaptionsoff
  \newpage
\fi



\bibliographystyle{IEEEtran}
\bibliography{references}
%



%

\begin{IEEEbiography}[{\includegraphics[width=1.05
in,height=1.5in,clip,keepaspectratio]{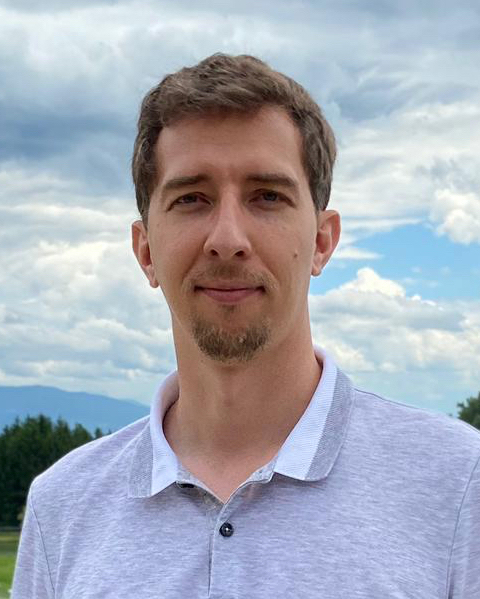}}]{Sven Kreiss}
  is a postdoc at the Visual Intelligence for Transportation (VITA) lab
  at EPFL in Switzerland focusing on perception with composite fields.
  Before returning to academia, he was the Senior Data Scientist at Sidewalk
  Labs~(Alphabet, Google sister) and worked on geospatial machine learning
  for urban environments. Prior to his industry experience, Sven developed
  statistical tools and methods used in particle physics research.
\end{IEEEbiography}

\begin{IEEEbiography}[{\includegraphics[width=1.05in,height=1.5in,clip,keepaspectratio]{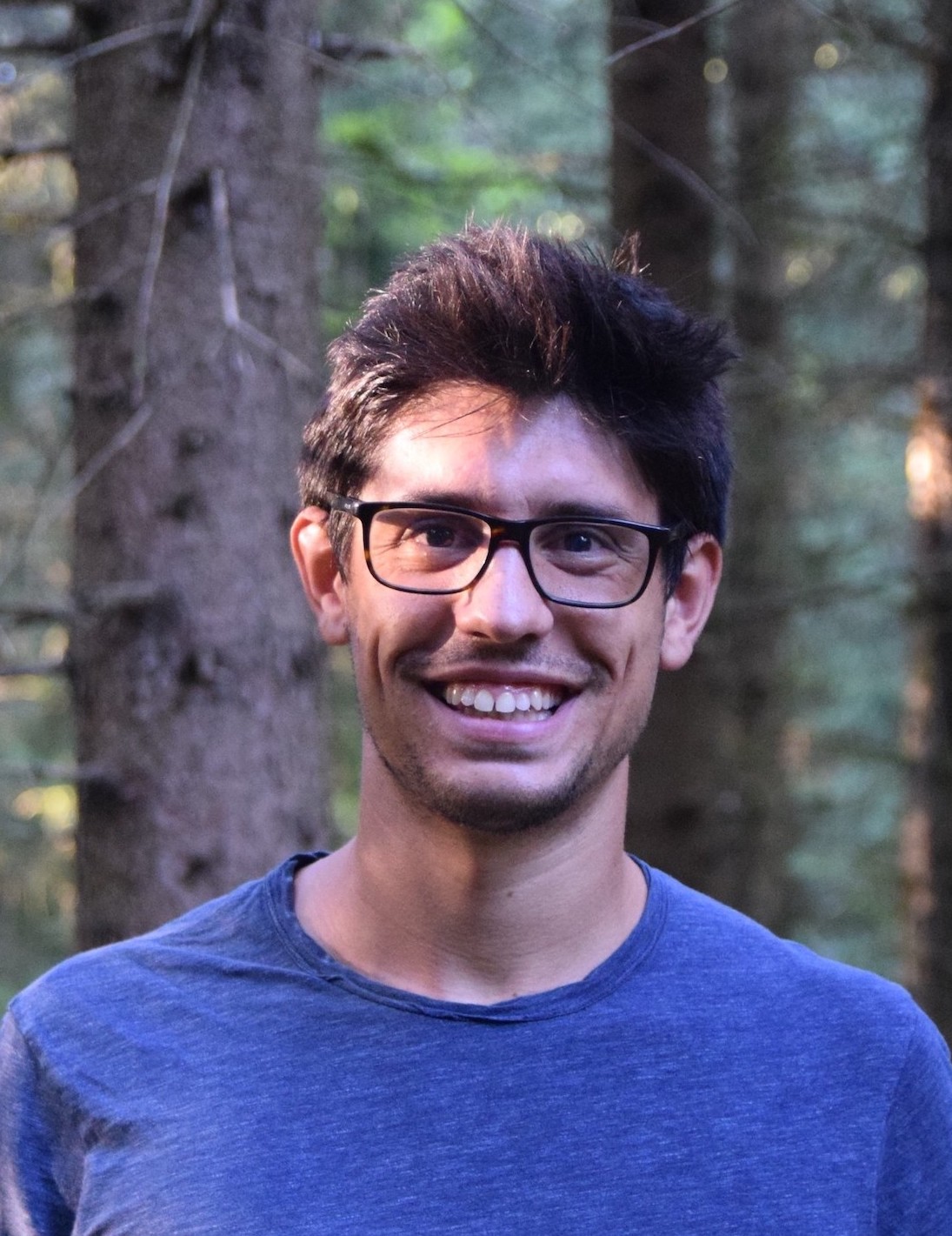}}]{Lorenzo Bertoni}
  is a doctoral student at the Visual Intelligence for Transportation (VITA)
  lab at EPFL in Switzerland focusing on 3D vision for vulnerable road users.
  Before joining EPFL, Lorenzo was a management consultant at Oliver Wyman
  and a visiting researcher at the University of California, Berkeley,
  working on predictive control of autonomous vehicles. Lorenzo received
  Bachelors and Masters Degrees in Engineering from the Polytechnic
  University of Turin and the University of Illinois at Chicago.
\end{IEEEbiography}

\begin{IEEEbiography}[{\includegraphics[width=1.05in,height=1.25in,clip,keepaspectratio]{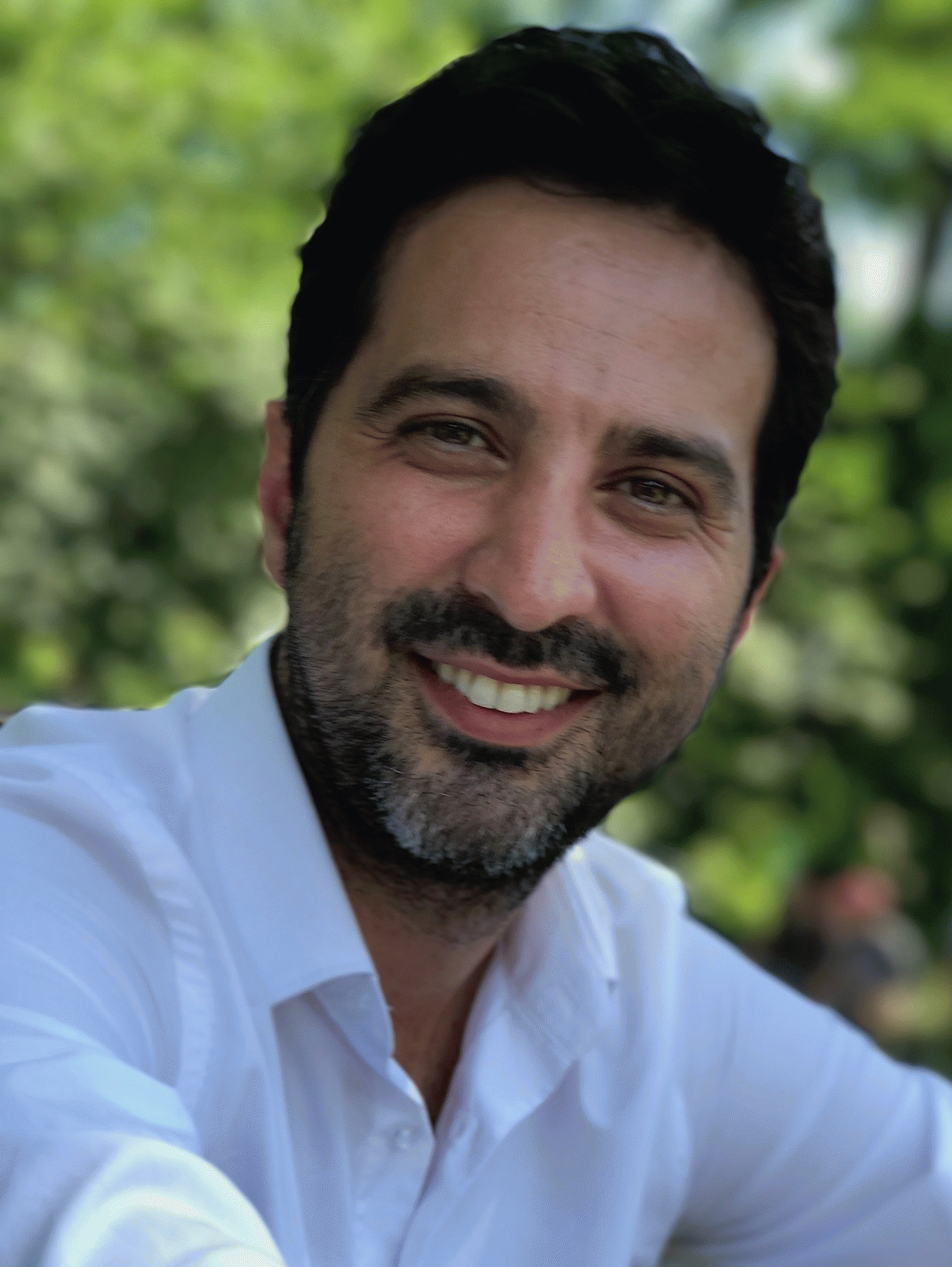}}]{Alexandre Alahi}
  is an Assistant Professor at EPFL. He spent five years at Stanford
  University as a Post-doc and Research Scientist after obtaining his Ph.D.
  from EPFL. His research enables machines to perceive the world and make
  decisions in the context of transportation problems and smart environments.
  He has worked on the theoretical challenges and practical applications of
  socially-aware Artificial Intelligence, i.e., systems equipped with
  perception and social intelligence. He was awarded the Swiss NSF early and
  advanced researcher grants for his work on predicting human social
  behavior. Alexandre has also co-founded multiple startups such as
  Visiosafe, and won several startup competitions. He was elected as one of
  the Top 20 Swiss Venture leaders in 2010.
\end{IEEEbiography}


\vfill


\end{document}